\documentclass[sigplan,screen]{acmart}

\usepackage{xcolor}
\usepackage{amsmath}
\usepackage{enumitem}  
\usepackage{tabularx}
\usepackage{adjustbox}
\usepackage{multirow}
\usepackage{caption}
\usepackage{booktabs}
\usepackage{array}
\usepackage{amsthm}
\renewcommand{\arrayrulewidth}{0.15pt}

\theoremstyle{definition}
\newtheorem{definition}{Definition}
\makeatletter
\newcommand\Vcline[1]{%
  \noalign{\vskip\arrayrulewidth\global\let\CT@do@color\relax}%
  \cline{#1}%
  \noalign{\vskip\arrayrulewidth}
}
\makeatother

\usepackage{adjustbox}
\usepackage{booktabs}
\usepackage{lipsum}         
\usepackage{booktabs}       
\usepackage{adjustbox}      
\usepackage{changepage}     
\usepackage{graphicx} 
\usepackage{physics}

\usepackage{svg}

\setcopyright{none}
\renewcommand\footnotetextcopyrightpermission[1]{}

\AtBeginDocument{%
  \providecommand\BibTeX{{%
    \normalfont B\kern-0.5em{\scshape i\kern-0.25em b}\kern-0.8em\TeX}}}

\begin{document}

\title{
Anomaly Detection  in Dynamic Graphs: A Comprehensive Survey}

\author{Ocheme Anthony Ekle}
\email{oaekle42@tntech.edu}
\orcid{https://orcid.org/0009-0003-6204-0657}
\affiliation{%
  \institution{Tennessee Technological University}
  \streetaddress{1020 Stadium Drive, 406.}
  \city{Cookeville}
  \state{TN}
  \country{USA}
  \postcode{38505}
}

\author{William Eberle }
\email{weberle@tntech.edu}
\orcid{https://orcid.org/0009-0009-1303-6102}
\affiliation{%
  \institution{Tennessee Technological University}
  \city{Cookeville}
  \state{TN}
  \country{USA}
  \email{weberle@tntech.edu}
}

\renewcommand{\shortauthors}{O.A Ekle and Eberle, et al.}


\begin{abstract}
  This survey paper presents a comprehensive and conceptual overview of anomaly detection using dynamic graphs. We focus on existing graph-based anomaly detection (AD) techniques and their applications to dynamic networks. The contributions of this survey paper include the following: i) a comparative study of existing surveys on anomaly detection; ii) a \textbf{D}ynamic \textbf{G}raph-based \textbf{A}nomaly \textbf{D}etection (\textbf{DGAD}) review framework in which approaches for detecting anomalies in dynamic graphs are grouped based on traditional machine-learning models, matrix transformations, probabilistic approaches, and deep-learning approaches; iii) a discussion of graphically representing both discrete and dynamic networks; and iv) a discussion of the advantages of graph-based techniques for capturing the relational structure and complex interactions in dynamic graph data. Finally, this work identifies the potential challenges and future directions for detecting anomalies in dynamic networks. This \textbf{DGAD} survey approach aims to provide a valuable resource for researchers and practitioners by summarizing the strengths and limitations of each approach, highlighting current research trends, and identifying open challenges. In doing so, it can guide future research efforts and promote advancements in anomaly detection in dynamic graphs.
\end{abstract}




\maketitle
\pagestyle{plain}
\section{Introduction}
\label{sec:introduction}

Anomaly detection involves identifying patterns in data that deviate significantly from a well-defined notion of normal behavior \cite{077_sp_chandola2009anomalyDetection}. In the works of Pang et al. \cite{084b_pang2019_anomaly_detection_with_deviation}, an anomaly is defined as a data point that deviates from the majority of other data points. These anomalies can manifest as patterns, observations, or data points that do not conform to the typical patterns observed in data. Anomaly Detection is an important task in both static and dynamic networks or graphs. Unlike static networks, where the topology remains constant, dynamic networks are constantly (or periodically) changing their node entities and edges \cite{028_sp_ranshous2015anomaly_detection_dynamic}. Dynamic networks have the ability to capture the temporal evolution of relationships in graphs, such as the insertion and deletion of nodes \cite{027_AnomRank_yoon2019fast}, the insertion and deletion of edges \cite{026_AddGraph_zheng2019addgraph}, and sudden pattern changes of sub-graphs or graph cliques \cite{025_alsentzer2020subgraph_SUBGNN, 042_yuan2023_motif}.

\begin{figure*}[htbp]
  \centering
  {\includegraphics[width=0.7\linewidth]{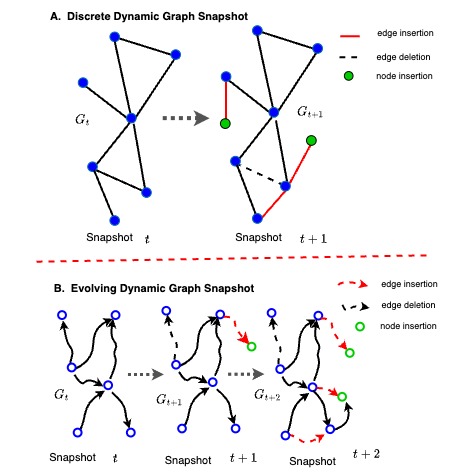}}

  \caption{  \textbf{Dynamic Graph Representation: }(\textbf{1a}) illustrates change in the dynamic graph $\mathcal{G} = (G_1, G_2, \dots, G_T) $, in which changes occur in distinct time intervals (that is, changes are not continuous but rather at specific time points, signifying a pattern of discrete changes over time).  (\textbf{1b}) captures a snapshot of an evolving dynamic graph ($\mathcal{G} = (V_t, E_t, \mathcal{T})$), where $V_t$ represents the node set, $E_t$ signifies the evolving edge set, and $\mathcal{T}$ denotes the sequence of time steps over which the dynamic graph evolves. The illustration is embedded within a continuous temporal context, reflecting changes that are not confined to specific time points but instead manifest as continuous transformations. }
  \label{fig1:dynamic_representation}
\end{figure*}

More interestingly, in the work of Michail et al. \cite{084_michail2018elements_dynamic_networks}, modern dynamic networks have proven to exhibit additional structural and algorithmic properties that go beyond the simple generalization of graphs . Yet, while the challenges of modeling dynamic networks as dynamic graphs are greater than on static graphs, the advantages of a dynamic representation are important.  
In Figure ~\ref{fig1:dynamic_representation}, the dynamic graph $\mathcal{G}$ is shown, illustrating two possible forms of graph representation at each time step. \textbf{(1a)} illustrates discrete dynamic changes occurring over distinct time intervals $\mathcal{G} = (G_1, G_2, \dots, G_T)$, while \textbf{(1b)} presents a snapshot of the evolving dynamic graph ($\mathcal{G} = (V_t, E_t, \mathcal{T})$), embedding changes through continuous evolution in transitioning graph streams, with $V_t$ as the node sets, $E_t$ representing evolving edges, and $\mathcal{T}$ indicating the sequence of time steps.


Graph algorithms have significant real-life applications in areas such as drug discovery \cite{084c_gaudelet2021_drug_discovery}, distributed systems \cite{084g_md2021_graph_distributed_sys}, IoT \cite{084f_chen2021_graph_transformer_IoT}, protein design \cite{084d_ingraham2019_graph_protein_design}, fraud detection \cite{022_zhang2022e-commerce_fraud_detection_graph}, social network analysis \cite{084e_fan2019_graph_social_analysis}, power grids \cite{030_li2021_DYNWATCH_power_grid}, and so on. Each of these domains has the property of being represented as a dynamic network. However, one of the predominant challenges in dynamic networks is detecting anomalous patterns \cite{028_sp_ranshous2015anomaly_detection_dynamic}, including nodes, edges, subgraphs, motifs, clusters, etc.

Early research on graph-based anomaly detection heavily relied on domain-specific knowledge and the utilization of statistical techniques \cite{019_sp_kim2022_AD_with_GNN}. Existing anomaly detection methods include density-based local outlier factor (LOF) \cite{AD1_2000_LOF, AD2_na2018_DILOF, AD3_salehi2016_fast_LOF}, based on tree structures and detecting anomalies as randomly separated points, and Isolation Forest (IF) \cite{AD4_liu2008_isolation_Forest, AD5_hariri2019_extended_Isolated_Forest}. Other anomaly detection approaches encompass traditional and similarity-based methods. Examples include the Reachable Distance Function for KNN classification \cite{28r_zhang2022_reachable_KNN1}, Quantum KNN for neighbor selection \cite{29r_li2023quantum_KNN2}, and Semi-Supervised learning for semantic similarities of clusters \cite{30r_zhou2023hypermatch3}.

However, Isolation Forest methods use tree-based structures defined by the maximum depth parameter and the sliding window's size, and they have constraints in capturing anomalies in long-range dependence that occur in streaming graphs. Deep learning methods have also been proposed for anomaly detection problems, such as autoencoders \cite{AD6_gong2019_Deep_AutoEncoder_unsuper, AD7_zong2018_deep_AutoEncoder_Gaussian_Mix, AD8_zhou2017_Robust_deep_autoEncoder}, generative adversarial networks (GANs) \cite{AD9_yang2021_GAN_anomaly_detection, AD10_bashar2020_time_series_AD_GAN}, and Recurrent Neural network (RNN)-based approaches \cite{AD11_saurav2018_RNN_Anomaly_detection, AD12_su2019_robust_RNN_Anomaly_Detection}. In recent years, researchers have also explored new emerging techniques for graph representation learning, such as reinforcement learning \cite{24r_bikos2021_reinforcement_1}, federated learning \cite{23r_pei2022personalized_federated_learning_1}, and quantum computing approaches \cite{20r_akbar2020towards_Quantum_Fruad_1, 21r_kukliansky2024network_AD_Quantum_2, 22r_rosenhahn2024_quantum_flowAD_3}. However, it is crucial to note that each method comes with varying challenges, such as speed, computational complexity, and scalability.

Despite the advancements made with deep learning, modern deep-learning frameworks are limited in their ability to process streaming data, and they are tailored to handle data in the form of sequences, images, and grid data. \cite{083_li_jure2022_power_of_GNN, 082_liu2020towards_Deep_GNN}. Despite their success, they are susceptible to over-fitting and inductive errors when dealing with a large number of parameters \cite{083a_yarotsky2017error_bounds_DNN, 083b_liang2017out_of_distri_NN}.

Graph-based algorithms have demonstrated their effectiveness in capturing changing relationships and dynamic changes between different structural parts and features through embedding and model learning by mapping how nodes and edges are connected \cite{033_sp_pazho2022survey_Graph_distributed_sys}. Graphs are well-suited for dynamic network tasks due to their inherent ability to effectively model interconnected relationships and patterns among entities (such as nodes, edges, and cliques) in both static and dynamic graphs. \\

\textbf{Why use a GNN for Anomaly Detection?} 

Graph-based approaches are considered for several reasons, including but not limited to the following:
\begin{enumerate}[label= \roman*. ]

    \item \textbf{Adaptation to Topological Structure:} In anomaly detection, the topological structure of dynamic graphs is important. Graph-based architectures, as exemplified by ST-GCN \cite{078_ST-GCN_yan2018spatial_temp_GNN}, excel in this regard. They dynamically propagate information through neighboring nodes, a feature particularly valuable when the network's structure evolves. This adaptability stands in contrast to other deep learning architectures such as CNN \cite{14_gehring2017cnn_seq}, LSTM \cite{hochreiter1997lstm}, GAN \cite{GAN_goodfellow2020generative} and ResNet \cite{ResNet_he2016deep}, which often assume fixed inputs like matrices or sequences.

    \item  \textbf{Integration of Multimodal Data:} GNNs have demonstrated the capability to integrate multimodal data with varying cardinalities and shapes, as proposed in MGNN by Gao et al. in 2020 \cite{gao2020_Multimodal_GNN}. This versatility is crucial for handling diverse and complex information, often encountered in real-life dynamic graph scenarios. Therefore, GNNs and graph-based models are most suitable for enhancing the efficacy of anomaly detection across multimodal data types.
   \item  \textbf{Scalability on Large-scale Networks:} Deep GNNs have shown good scalability on large-scale networks and complex graph representations. This was demonstrated in GNNautoScale by Fey et. al. in 2021 \cite{079_fey2021GNNautoScale}. GNNautoScale leverages the localized message-passing algorithms to prune entire sub-trees of the computation graph by utilizing historical embeddings from prior training iterations, leading to constant GPU memory consumption with respect to input node size without dropping any data. Deep GNNs have proven effective in scaling to handle the intricacies of dynamic networks, a key aspect in anomaly detection for real-world graph networks.
   
   \item  \textbf{Efficiency in Model Interpretability:} Graph-based approaches demonstrate high efficiency in deep model interpretability and explainability. Recent contributions include the works of F{\"u}{\ss}l et al. in 2022 \cite{fussl2022interpretability_KGraph} on the interpretability of knowledge graphs and the interpretable learning of dynamic graph-convolutional networks (GCNN) by Zhu et al. in 2022 \cite{zhu2022interpretable}. Graph-based interpretability has become particularly relevant in anomaly detection tasks, where understanding and interpreting the model's decisions is crucial for practical deployment and decision-making.
    
    \item \textbf{Incorporation of Self-Attention and Transformers:} Recent studies have indicated that GNNs can be integrated with self-attention \cite{22_shaw2018self_atten_relative_postion} and transformers \cite{17_vaswani2017attention_is_all_you_need} as special cases, highlighting their adaptability. Examples include GAT by Velickovic et al.\cite{064_0velickovic2017_GAT} in 2017, Graph-Transformer by Yun et al. \cite{080_yun2019graph_Transformer_paper} in 2017, and Graphomer by Ying et al. \cite{063_ying2106_Graphomer_Do_transformers_Bad_Graph} in 2021 (detailed explanations are provided in Section \ref{sec5:Ad_approaches} and Table \ref{tab2_AD_approach}). The adaptability of the Graph Transformer is crucial for extending existing deep-learning architectures for anomaly detection in dynamic graphs. GNNs can capture intricate patterns and complex graph dependencies in dynamic networks, thereby enhancing their anomaly detection capabilities.

\end{enumerate}

\subsection{Existing Surveys on Anomaly Detection}

Previous survey studies done on anomaly detection (AD) have dealt with the subject from different perspectives. Chandola et al. \cite{077_sp_chandola2009anomalyDetection} provides a structural review on Anomaly detection, \cite{2a_sp_pang2021deep_learning_AD, 2b_sp_pang2021deep_learing_AD_tutorial, 2d_sp_chalapathy2019deep_learning_AD_most_cited} focused on deep learning approaches. \cite{2e_sp_ruff2021_deep_and_shallow_AD} conducts a unifying review that connects traditional shallow and deep learning methods. \cite{2h_sp_thudumu2020_AD_for_high_dim_big_data, 2g_sp_habeeb2019real_time_data_data_AD} focuses on anomaly detection on real-time big data. Some survey works capture specific domains such as fake news detection in social networks \cite{2i_sp_ahmed2022_fake_news_survey}, financial domains \cite{2j_sp_ahmed2016_financial_domain_survey}, IoT and sensor networks \cite{032_sp_kyle2023survey_AD_IoT_Sensor_Networks}, distributed systems \cite{033_sp_pazho2022survey_Graph_distributed_sys}, time series \cite{034_sp_ho2023graph_based_AD_time_series}, etc. A recent study \cite{2c_sp_wang2021survey_Attributed_graph} focuses on attributed graph queries, while \cite{001_sp_ding2022_data_augmentation_Deep_graph} conducts a review on
data augmentation approaches for deep graph representation learning,  and \cite{04_sp_waikhom2021GNN_Methos_shallow_deep_embedding} aims to classify graph-based semi-supervised learning techniques based on their embedding methods (shallow graph embedding and deep graph embedding). \cite{4a_zhou2020_GNN_method_application} provides a broad pipeline of graph neural networks (GNNs) and discusses the variants of each module. The survey also presents research on both theoretical and empirical analyses of GNN architectures. 

Despite the increasing number of surveys targeted at graph-based AD, most techniques are focused on GNN models and anomaly techniques in static graphs. Ranshous et al. \cite{028_sp_ranshous2015anomaly_detection_dynamic} provides one of the first surveys on anomaly detection in dynamic graphs. The article gives a broad overview on data mining in dynamic networks by introducing four common variants of anomalies associated with dynamic networks namely, node, edge, subgraph, and event-level anomalies. The authors in \cite{028_sp_ranshous2015anomaly_detection_dynamic} further categorize graph-based models into five primary groups, originating from the underlying ideas behind each approach: communities, compression, decomposition, distance, and probabilistic model-based methods. 

The authors in \cite{3a_sp_holme2012_temporal_network, 3b_sp_holme2015modern_temporal_network_survey} provide an introductory review of temporal networks, including an overview of methods, modeled entities, and challenges associated with temporal networks, but with no detailed analysis of the dynamic network evolution or algorithms pertaining to real-world systems. Kazemi et al.
\cite{3c_sp_kazemi2020representation_dynamic_graph} present a theoretical approach to representation learning for dynamic graphs, with a focus on time-dependent embedding techniques designed to capture the fundamental characteristics of nodes and edges within evolving graphs.  

Most recently, Skarding et al. \cite{071_sp_skarding2021_foundations_dynamic_graph_survey} provides an outline of dynamic network models using GNNs. In this work, the authors categorize dynamic models into statistical, stochastic actor-oriented, and dynamic network representation learning. Furthermore, the authors in \cite{071_sp_skarding2021_foundations_dynamic_graph_survey} captures the deep learning approaches for encoding a dynamic topology and an overview of an encode-decoder framework in dynamic link prediction. However, the techniques outlined in \cite{028_sp_ranshous2015anomaly_detection_dynamic, 071_sp_skarding2021_foundations_dynamic_graph_survey} do not provide separate explanations for each learning setting on static and dynamic graphs. 

In contrast to the existing works, our study offers a more comprehensive survey of the current frameworks used for anomaly detection (AD) in dynamic graphs. We organize the current methods for AD in dynamic graphs into four categories: traditional machine-learning models, matrix transformations, probabilistic approaches, and deep-learning approaches. We also provide a chronological timeline of these dynamic graph models. 

Furthermore, we present an in-depth discussion of the different ways dynamic graphs are being represented in real-world data and a highlight of the current datasets and metrics used in the literature.


\begin{table*}
  \caption{A Comparison Between Existing Surveys on Anomaly Detection. We mark edge and sub-graph detection as in our survey because we review more deep learning based works than any previous surveys.}
  \label{tab1_survey_table}
  \begin{adjustbox}{center}
  \resizebox{\textwidth}{!}{
  \begin{tabular}{l|l|l|l|lll|l}
    \toprule
    \multirow{2}{*}{\textbf{Surveys}} &\multirow{2}{*}{\textbf{Year}} & 
    \multirow{2}{*}{\textbf{Research Emphasis and Description}} 
     & \multirow{2}{*}{\textbf{Graph}}  & \multicolumn{3}{c|}{\textbf{Graph Level Tasks}} & \multirow{2}{*}{\textbf{Network}} \\
   \cmidrule(lr{0.15pt}){5-7}
    & &   & \textbf{Based} & \textbf{Node} & \textbf{Edge} & \textbf{Subgraph} & \textbf{Types} \\
    \midrule
    Chandola et al.  \cite{077_sp_chandola2009anomalyDetection}& 2009  & Traditional AD techniques &  - & - & - & - &  - \\
     Holme et al. \cite{3a_sp_holme2012_temporal_network} & 2012 &  A study of temporal networks and dynamic graphs  &  \checkmark & - & - & - & Dynamic\\
    Holme et al. \cite{3b_sp_holme2015modern_temporal_network_survey} & 2015 & Methods for analyzing and modeling temporal networks. &  - & - & - & - & Dynamic\\
      Ranshous et al \cite{028_sp_ranshous2015anomaly_detection_dynamic} & 2015 &In-depth review of Dynamic Graph AD up to 2015.  &  \checkmark & \checkmark & \checkmark & \checkmark & Dynamic\\
    Ahmed et al. \cite{2j_sp_ahmed2016_financial_domain_survey} & 2016 & Clustering based AD methods in Financial Domain  &  - & - & - &  - & -\\
    Yu et al. \cite{2k_sp_yu2016_social_media_AD_survey} & 2016  & Social Media AD techniques   &  \checkmark & - & - & - & Static \\
    Habeeb et al. \cite{2g_sp_habeeb2019real_time_data_data_AD} & 2019& Real-time big data for AD  &  - & - & - & - &  -\\
     Thudumu et al. \cite{2h_sp_thudumu2020_AD_for_high_dim_big_data} & 2020 & AD techniques for Big data  &  - & - & - & -  & -\\
    Zhou et al. \cite{4a_zhou2020_GNN_method_application} & 2020& A review of DL \& GNN for graph learning tasks  &  \checkmark & - & - & - & -\\
        
    Kazemi et al. \cite{3c_sp_kazemi2020representation_dynamic_graph} & 2020 & Embedding techniques for dynamic graph representation.  &  \checkmark & - & - & - & Dynamic\\
    
    Pang et al. \cite{2a_sp_pang2021deep_learning_AD}  & 2021 & DL for Anomaly Detection  &  - & - & - & - &  -\\
    Pang et al. \cite{2b_sp_pang2021deep_learing_AD_tutorial} &2021 &  Deep AD techniques  &  - & - & - & - &  -\\
      Wang et al. \cite{2c_sp_wang2021survey_Attributed_graph} & 2021 & Survey of attributed graph queries  &  \checkmark & - & \checkmark & \checkmark & Static\\
      Ruff et al. \cite{2e_sp_ruff2021_deep_and_shallow_AD} & 2021  & Deep and shallow learning for AD  &  - & - & - & - & -\\

    Waikhom et al. \cite{04_sp_waikhom2021GNN_Methos_shallow_deep_embedding} & 2021  & Survey on GNN models, applications, and learning techniques.  &  \checkmark & - & - & - & Static\\
          
     Skarding et al. \cite{071_sp_skarding2021_foundations_dynamic_graph_survey} & 2021 &  
            GNN \& DGNN techniques for dynamic graph.   &  \checkmark & - & \checkmark & - & Dynamic\\
            
      Ahmed et al. \cite{2i_sp_ahmed2022_fake_news_survey} & 2022 & AD techniques for Social Network Fake News   &  - & - & - &  - & -\\
      
        Pazho et al. \cite{033_sp_pazho2022survey_Graph_distributed_sys} & 2022 &   DAD \& Graph-based AD in Distributed systems   &  \checkmark  & - & \checkmark  &  - & Dynamic\\
         
          Ding et al. \cite{001_ding2022Data_Augmentation} & 2022 & GNN \& Data augmentation for deep graph learning   &  \checkmark & - & - & - & Static\\
          Ho et al. \cite{034_sp_ho2023graph_based_AD_time_series} & 2023  & DAD \& Time Series AD techniques   &  \checkmark & \checkmark  & \checkmark  & \checkmark & -\\
          DeMedeiros et al. \cite{032_sp_kyle2023survey_AD_IoT_Sensor_Networks} & 2023 & Deep AD techniques in IoT and Sensor Networks   &  \checkmark & - & - & - &  Both\\
    \hline
     \textbf{Our Survey (DGAD)} & \textbf{2023} & \textbf{ Current AD techniques for Dynamic Graph (2016 -2023) } & \checkmark & \checkmark  & \checkmark & \checkmark &  Dynamic\\
    \bottomrule
    \multicolumn{8}{l}{*\textbf{AD}: Anomaly Detection, \textbf{DL}: Deep Learning, \textbf{DAD}: Deep Anomaly Detection, \textbf{DGAD: } Dynamic Graph-Based Anomaly Detection  } \\
    \multicolumn{8}{l}{*\textbf{GADL}: Graph Anomaly Detection with Deep Learning, \textbf{TSAD:} Time Series Anomaly Detection  Learning, } 
    
  \end{tabular}
  }
   \end{adjustbox}
\end{table*}

\subsection{Proposed Framework and Structure}

In Table \ref{tab1_survey_table}, we present a comparative outline of existing surveys on anomaly detection. The table outlines the method utilized in each study and highlights common trends. The focus areas of these surveys are also examined, distinguishing between static and dynamic contexts, and their coverage across different domains is noted. Notably, the inclusivity of specific anomaly patterns in graphs, such as nodes, edges, and sub-graph levels, is emphasized.

\begin{figure*}[h]
  \centering
  {\includegraphics[width=0.6\linewidth]{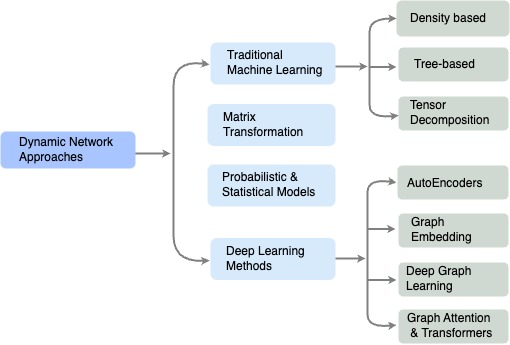}}
  
  \caption{ An Overview of Survey Framework on \textbf{D}ynamic \textbf{G}raph-based \textbf{A}nomaly \textbf{D}etection (\textbf{DGAD}). }  
  \label{fig2:survey_framework}
\end{figure*}

We aim to provide an overview of our survey categorization approach for anomaly detection in dynamic graphs. This is illustrated in Figure  ~\ref{fig2:survey_framework}. We will provide a more detailed explanation of each individual component of our framework in Section ~\ref{sec5:Ad_approaches}.

\subsection{Contributions}

In summary, the contributions of this survey are as follows:
\begin{itemize}
    \item First, we provide a high-level overview of existing surveys on anomaly detection, graph data mining, and graph representation learning, as presented in Table  \ref{tab1_survey_table}.
    
    \item We then introduce the concept of anomalies and discussed the three main types: point, contextual, and collective anomalies. After that, we provide a detailed mathematical definition of anomalies in both static graphs (including some recent works) and dynamic graphs. This definition includes tasks at the node, edge, and sub-graph levels, which are the most common graph-based tasks found in the literature.
    
    \item We further provide an extensive overview of the existing techniques and methods used for detecting anomalies in dynamic graphs, along with a comparative analysis of these methods. This is illustrated in Figure ~\ref{fig2:survey_framework} and in a detailed summary in Table ~\ref{tab2_AD_approach}.

    \item Additionally, we discuss the representation of dynamic graph patterns in data, covering both discrete and continuous graphs. See the summary in Table   ~\ref{tab:dynamic_rep_comparison}.
    \item Finally, we discuss the potential challenges and future directions in dynamic graph anomaly detection.

\end{itemize}

Given the current growth in graph-based research, our \textbf{DGAD} survey approach is anticipated to be valuable to the graph-learning research community by providing valuable resources for researchers and practitioners by summarizing the strengths and limitations of each approach, highlighting current research trends, and identifying open challenges.

The rest of the paper is organized as follows: Section \ref{sec2:backgroud} provides a background study and important definitions of terminologies. In Section \ref{sec3:GNN_intro}, we give an overview of the architecture of graph neural networks (GNNs). Section \ref{sec4:dynamic_graph_rep} discusses the different representations of dynamic networks. Our survey approach, DGAD methods, and application are presented in Section \ref{sec5:Ad_approaches}. In Section \ref{sec6:dataset_metrics}, we present datasets and evaluation metrics used in the literature. Finally, in Section \ref{sec7:future_works}, we discuss the comparison, challenges, current trends, and future directions of dynamic graph research.


\section{Background}
\label{sec2:backgroud}

In this section, we introduce the fundamental concept of anomalies, beginning with an exploration of their types and definitions. Subsequently, we defined anomalies in both static and dynamic graphs, paving the way for a comprehensive understanding of anomalies in dynamic networks. 

To enhance clarity, we will use the terms ``\textit{graphs}'' and ``\textit{networks}'' interchangeably throughout this discussion.

\subsection{Understanding Anomalies}
\label{sec:understand_AD}
Anomalies in real-world data manifest infrequently and come in diverse forms and structures. They tend to be domain-specific, encompassing anomalies like irregular intrusion patterns, fraudulent transactions, outliers in social networks, and the detection of unusual patterns in industrial machinery, among others. In line with existing survey by Chandola et al. \cite{077_sp_chandola2009anomalyDetection}, we categorize anomalies into three main types: point anomalies, contextual anomalies, and collective anomalies, taking into account their characteristics and prevalence in the data.

However, in \ref{sub_sec2:AD_in_graph}, we will elaborate on various categories of anomalies, specifically within graph data.

\subsubsection{Point Anomalies} Also referred to as individual or atomic anomalies, \textit{point anomalies} represent data points that stand out from the typical or expected data distribution within a dataset. These anomalies are typically isolated instances and do not depend on the surrounding dataset. Detecting point anomalies often involves applying statistical methods and predefined thresholds.

Among the common algorithms employed for point anomaly detection in the literature is the Z-score (standard score) \cite{b1_AD_DBSCAN2011_pointAD}, which measures the degree of standard deviations a data point deviates from the dataset's mean. Additionally, density-based clustering algorithms are valuable for grouping data points into individual clusters; this is shown in the works of Almuzaini et al. \cite{b2_almuzaini2010_Density_pointAD}. In real-world situations, point anomalies could be found in different domains, such as unusual fraudulent transactions \cite{022_zhang2022e-commerce_fraud_detection_graph}, unexpected data packets in a network traffics such as F-FADE \cite{043_F-FADE_chang2021f},  DYNWATCH \cite{030_li2021_DYNWATCH_power_grid}, MIDAS \cite{040_bhatia2022_MIDAS_latest} and many others,  abnormal medical diagnostic results \cite{12r_ahmedt2021_Graph_Medical}, abrupt changes in stock market prices \cite{s5_hooi2017_graph_based_fraud_detection}, etc.

\subsubsection{Contextual Anomalies}

Also known as \textit{conditional anomalies}, in this scenario, a data point exhibits an anomaly within a particular context. The idea of context derives from the inherent structure of the dataset. For instance, examining network traffic patterns based on time, where an abrupt increase in multiple login attempts occurs during non-business hours. Another example of a contextual anomaly might arise if network traffic suddenly originates from an unusual geographic location outside the company's usual operational coverage, among other scenarios.

Commonly used techniques for detecting contextual anomalies include statistical models \cite{b3_song2007conditional_AD}, rule-based methods \cite{b4_roy2010rule_based_AD}, Gaussian threshold-based models \cite{b5_wang2011_Stat_threshold_AD}, and machine learning, just to name a few.

\subsubsection{Collective Anomalies}
Also referred to as group anomalies or global anomalies, \textit{collective anomalies } occurs when a subset of data instances or groups exhibits unusual patterns that deviate from the entire data set. These kinds of anomalies focus on the collective behavior of groups rather than isolated data points \cite{077_sp_chandola2009anomalyDetection}.  Real-world examples can be found in coordinated distributed denial of service (DDoS) attacks, where multiple computers collectively exhibit malicious behavior. Another instance can be found in the work of Zhang et al. (\cite{13r_zhang2024bayesian_Bot_attack}), who used Bayesian graph local extrema convolution for bot detection in 2024. Their work addressed the detection of coordinated bot attacks on social media platforms, specifically those aimed at propagating fake news or engaging in online manipulation.

Frequently used methods in the literature, including clustering-based \cite{b6_wang2022_clustering-based}, density-based \cite{AD2_na2018_DILOF, AD3_salehi2016_fast_LOF}, matrix factorization methods \cite{mf1_sun2006_dynamic_Tensor, mf3_teng2017_multi_view_Time_series, mf4_yu2017_Tem_Fact_Network}, and graph-based algorithms \cite{030_li2021_DYNWATCH_power_grid, 036_fang2023_AER_anonymous_edge, 041_gao2023_GDN_structural_shift}, have proven effective in recent research.

\subsection{Anomaly Detection in Static Graph}
\label{sec2.2:static_graph}
In Section \ref{sec:introduction}, we introduced the concepts of static graphs, which depict networks with a fixed structure, and dynamic graphs, which evolve over time. Here, we present the mathematical formulation for static graphs and the definition of anomalies in static networks. We will also briefly review a handful of pieces of literature in this area; however, it's worth noting that our survey does not focus on anomaly detection in static networks.

It is imperative to establish a more formal definition of the term ``\textit{Graph}'' to provide a precise conceptual foundation.

\begin{definition}
\label{def_1}

A graph ${G} = (\mathcal{V}, \mathcal{E})$ is defined as a pair consisting of a set of nodes $\mathcal{V}$ and a set of edges connecting these nodes $\mathcal{E}$. Here, $\mathcal{V} = \{v_1, v_2, \ldots, v_n\}$ represents the node set, where each $v_i$ represents an individual node or vertex within the graph. The set of edges, denoted by $\mathcal{E}$, is a subset of the Cartesian product of $\mathcal{V}$ with itself, i.e., $\mathcal{E} \subseteq \mathcal{V} \times \mathcal{V}$, and it defines the relationships or connections between the nodes.
\end{definition}
In a simple graph, we represent an edge from node $u \in \mathcal{V}$ to node $v \in \mathcal{V}$ as $(u,v) \in \mathcal{E}$. A graph, denoted as ${G}$, can be either undirected or directed in nature. In an \textbf{undirected} graph, an edge $(u, v)$ forms an unordered pair of vertices, and the relationships between nodes are symmetric. In contrast, in a \textbf{directed} graph (or digraph), each edge has a specific direction, creating asymmetric relationships between nodes.

\subsubsection{ What is an anomaly in a static graph?}  Anomalies in static networks could be classified by the anomalous entities that are spotted, such as nodes, edges, subgraphs, motifs, etc. We provide formal definitions for anomalies within a static network in Definition~\ref{def_2:AD_static} and Definition~\ref{def_3:AD_static_attribute}.

\begin{definition}[\textit{Node Anomaly in  static graph:}] 
\label{def_2:AD_static}
Given a static graph ${G = ({V}, {E})}$ from definition \ref{def_1}, with the node set ${V}$ and edge set ${E}$, we can define a \textbf{node-level anomaly} if there exists a node $u \in {G}$ and a statistical measure denoted as $\mathbb{N_\omega} (n) > \theta$, which exceeds a predefined threshold value $\theta$.

In simpler terms, $\mathbb{N_\omega} (n) > \theta$ indicates that node $\mathfrak{u}$ is anomalous in the graph ${G}$.
\end{definition}

Furthermore, static networks can also be modeled as a \textbf{static attributed graph} $G = (V, E, A_e)$, where ${V}$ is the set of nodes, $E$ is set of edges with $ e \in {E} : e = (u, v) \hspace{0.2cm} \forall u, v \in V$. $A_e$ is the set of attributes with edges, denoted as a function $A_e : E \rightarrow \mathbb{R}^n$, where $n$ is the number of attributes per edges. In other words, for all edges $e \in E$, the attribute matrix $ A_e (e)$ provides an attribute vector.

\begin{definition}[\textit{Edge Anomaly in Attributed static graph:}] \label{def_3:AD_static_attribute}
Given an attributed static graph $G = (V, E, A_e)$  as defined above, an edge $e \in E$ is considered anomalous with respect to its attributes if it exhibits a deviation from the expected attribute vector distribution of edges in the entire graph $G$. Let $\mathbb{M}(A_e(e))$ denote the statistical measure of deviation (e.g., standard deviation, Z-score) of a given edge $e$ from the normal distribution of edge attributes, and if this deviation exceeds a predefined threshold $\theta_e$, then $e$ is labeled as an anomalous edge.

\end{definition}

\subsubsection{Existing works in Static Graphs:} The majority of the proposed methods on graph data mining for unusual patterns have largely focused on modeling static networks \cite{083_you2022_Roland}. 

Among methods focusing on static graphs are CATCHSYNC \cite{s8_jiang2016_CATCHSYNC} by Jiang et al. on anomalous node detection. The model computes the node characteristic features of the graph, putting into consideration the node degree centrality and authoritative nodes (or Ego node), and subsequently identifies nodes whose neighbors exhibit close proximity in the feature space. In 2022, Liu et al. \cite{018_liu2022_BOND_static_attributed_Graph} proposed BOND, an unsupervised node detection approach for static attributed graphs. BOND aims to evaluate the performance of different GNN-based algorithms in detecting both structural and contextual anomalies. Zhao et al. \cite{020_zhao2021_PAMFUL_pattern_mining_feature_learning} in 2021 developed the PAMFUL framework, which synergistically combines pattern mining algorithms and feature learning via a GNN encoder for graph anomaly detection, effectively leveraging both local and global structural patterns. PAMFUL uses the GNN encoder to perform feature aggregation and leverages the Random Walk algorithm to capture the global pattern of the graph structure.

Wang et al. introduced the EGNN model \cite{021_gong2019_edge_features_GNN}, an Edge Feature in GNN that utilizes a Doubly Stochastic Edge Normalization instead of the symmetric normalization approaches used in GCN \cite{066_GCN_kipf2016semi_supervised_GCN} and GAT \cite{064_0velickovic2017_GAT}. By utilizing multi-dimensional positive-valued edge features, EGNN \cite{021_gong2019_edge_features_GNN} eliminates the challenges faced by GAT \cite{064_0velickovic2017_GAT}, which can only handle binary edge indicators, and the limitations of GCN \cite{066_GCN_kipf2016semi_supervised_GCN}, which can only handle one-dimensional edge features.

Other frameworks targeted at static graph data include FRAUDAR \cite{s5_hooi2017_graph_based_fraud_detection} by Hooi et al., a Graph-Based Fraud Detection in the Face of Camouflage. The FRAUDAR fraud detection algorithm incorporates the greedy algorithm and density-based metrics to detect both camouflaged and non-camouflaged fraud in real-world data. CATCHSYNC \cite{s6_jiang2016_CatchSync} by Jiang et al. is a graph mining approach, a parameter-free and scalable method for automatically detecting suspicious nodes in large directed graphs based on synchronized behavior and rare connectivity patterns. Zhang and Chen \cite{s3_zhang2018_link_prediction} focus on link prediction based on Graph Neural Networks (GNN), where they employ a heuristic approach involving the extraction of a local subgraph around each target link and learning a function to map the subgraph pattern to the existence of each link.

In 2021, You et al. \cite{s2_you2021_identity_aware} introduced the ID-GNN framework, a class of message passing techniques called Identity-aware Graph Neural Networks. ID-GNN extends existing GNN models by inductively incorporating nodes' identities into the message-passing process. When embedding a specific node, ID-GNN starts by extracting the ego (or authority) network centered around that node and subsequently conducts multiple rounds of heterogeneous message passing. Throughout this procedure, distinct sets of parameters are applied to the central node in contrast to the other nodes within the ego network. Similar to You et al.'s work \cite{s2_you2021_identity_aware}, Sengupta \cite{s1_sengupta2018_AD_static_egonet} developed an anomaly detection framework for static networks based on statistical inferences using the egonet method. This approach is effective for detecting anomalous cliques and subgraphs in static networks, with a two-step process: firstly, detecting the presence of a small anomalous clique, and secondly, identifying the node that forms the clique. Widely adopted graph representation techniques, such as DeepWalk \cite{perozzi2014_DeepWalk}, Node2Vec \cite{grover2016_Node2Vec}, and LINE \cite{tang2015_LINE}, have demonstrated their capability in generating node representations across graph networks and have been used to validate the performance of anomaly detection \cite{024_sp_ma2021comprehensive_survey_AD_Deeplearning}.

Recent and classical methods on anomaly detection on static graphs include SCALA \cite{01r_he2024_SCALA} by He et al., published in 2024, an unsupervised multi-view contrastive learning approach for anomaly detection in attributed networks. SCALA leverages the sparsification of networks, which filters the abnormal relationships based on the similarity between the nodes. This approach reduces the divergence on graph-level embedding caused by anomalous nodes. In 2024, Xu et al. introduced ADVANCE \cite{02r_xu2024unsupervised_ADVANCE}, a novel view-level unsupervised contrastive learning framework for detecting anomalies on an attributed static graph. The framework combines graph contractive learning-based and network reconstruction-based modules, improving anomaly detection efficiency through the joint optimization of these complementary components. This method offers strong assurance of the safety of consumer electronics. In 2023, Penghui et al. introduced LRAGAD \cite{03r_penghui2023lragad_LRAGAD}, a local information recognition system for attribute graph anomaly detection. LRAGAD employs anomaly score estimation to predict outliers based on the contextual structural information of the graph. Simultaneously, it utilizes a deep self-encoder to reconstruct both the structural and attribute information of the static attribute graph by generating various substructures from the target nodes.

In 2024, Jing et al. introduced SCN$\_$GNN \cite{04r_chen2024_SCN_GNN}, a Strongly Connected Nodes-Graph Neural Network designed for fraud detection. Their approach proposes two node sampling strategies, incorporating strong node information and graph topology information fusion. Specifically, it includes the Structured Similarity-Aware Module (SSAM) for up-sampling sparse graph nodes and the Strong Node Module (SNM) for down-sampling based on strong node information and original features. These techniques enhance the detection of neighboring nodes, adding value to the overall learning task.

Other classical anomaly detection approaches for static graphs include XGBOD \cite{05r_zhao2018xgbod}, Bayesian models, spectral analysis, and relational learning. Additionally, widely utilized standard graph neural networks (GNNs) such as GraphSAGE \cite{089_hamilton2017_GraphSAGE}, GAT \cite{060_huang2021_HO-GAT}, PNA \cite{07r_corso2020principal_PNA}, RGCN, and specialized GNNs like CARE-GNN \cite{08r_dou2020enhancing_CARE_GNN}, AMNet \cite{09r_chai2022can_AMNet}, and BWGNN \cite{10r_tang2022rethinking_BWGNN} have been prevalent in the field. For a more comprehensive exploration of static graph approaches, it is recommended to consult GADBench, a recent benchmark paper on supervised graph anomaly detection authored by Jainheng et al. \cite{11r_tang2024gadbench_Gadbench} in 2023.

While there exist several methods aimed at anomaly detection in static graphs across various domains, we have chosen only to spotlight a select few as real-world networks are dynamic in nature and constantly evolving, which is the primary focus of this work.

\subsection{Anomaly Detection in Dynamic Graph}

Dynamic graphs are frequently used to model real-world networks, capturing their ever-changing patterns and relationships. These changes can manifest through the detection or addition of nodes, edges, or subgraphs. In this section, we establish the mathematical formulation for dynamic graphs and present the types of anomalies that can be found in them.

It is worth noting that throughout our survey, we use the terms ``graph'' and ``network'' interchangeably to refer to the same evolving graph concept. However, it is important to acknowledge that in some literature, the term ``graph'' is more commonly used within the machine learning community, while ``network'' is historically used in data mining and network science. 

The full set of symbols and notations can be found in Table ~\ref{tab2:notations}. In subsection ~\ref{sub_sec2:AD_in_graph}, we start by introducing the three major anomaly tasks prevalent in the dynamic graphs literature: node, edge, and subgraph-level tasks, along with their corresponding mathematical representations.

\begin{definition}[\textit{Dynamic graph:}] \label{def_4:dynamic}
Given a graph $G = (V, E)$, where $V= \{ v_1, \dots, v_n \}$ is the node set and $E \subseteq V \times V$ is the edge multi-set, a \textbf{dynamic graph} $\mathcal{G} = \{ G_t \}_{t=1}^T$ can be defined as a sequence of ordered sets of graph snapshots at different time steps $t$, where $T$ is the total number of time steps. Each snapshot is considered as a static graph $G_t = (V_t, E_t \subseteq (V_t \times V_t) )$ with vertex set $V_t = \{ v\in V \| i_v = t \}$ and edge set $E_t = \{ e \in E \| i_e =t\}$, which may consist of plain or labeled edges.

\begin{table}[htbp]
    \caption{List of Notations}
    \label{tab2:notations}
    \centering
    \begin{tabular}{p{1.5cm} p{6cm}}
    \toprule
       \textbf{Symbol} & \textbf{Meaning} \\
    \toprule
        $G$    & a graph with a set of nodes ${V}$ and edges ${E}$. \\
        $V$    & represents the node set $\{v_1, v_2, \dots, v_n\}$, where $v_i$ are individual vertices. \\
        $E$    & the set of edges. \\
        $\mathcal{G}$ & denotes a dynamic graph. \\
        $G_t$  & a sequence of snapshots $\{ G_t \}_{t=1}^T$ at different time steps $t$. \\
        $T$    & the total number of time steps. \\
        $V_t$  & the vertex set for the graph at time point $t$. \\
        $f : V \rightarrow \mathbf{R}$ & a function that maps elements from the set of vertices $V$ to a real numbers $\mathbf{R}$
        \\
        $\forall v^{\prime} \in V^{\prime}$ & the condition applies to each vertex (or node) $v^{\prime}$ in the set $V^{\prime}$. \\
        $E_t$  & the edge set at time point $t$. \\
       
         $|f(v^{\prime}) - \hat{f}|$ & the absolute difference between the score assigned to vertex $v^{\prime}$ by the scoring function $f(v^{\prime})$ \\
        $\Phi_\omega$ & node-level anomalous scoring function  \\
        $\Phi_{e_{ij}(t)}$ & edge-level anomalous scoring function  \\
       \bottomrule
    \end{tabular}
\end{table}

\end{definition}
\subsection{Types of Anomalies  in Dynamic Graph}
\label{sub_sec2:AD_in_graph}
Anomalies can take on various forms within dynamic graphs due to the evolving and dynamic nature of network data. In this context, we will explore common types of anomalies, including node-level anomalies, edge anomalies, and sub-graph or clique anomalies.
\subsubsection{Node-level Anomalies:}
The goal of anomalous node (or vertex) detection is to identify a group of vertices or nodes where each vertex in this group exhibits a 'unique' or 'unusual' evolution when compared to the entire vertices within the graph \cite{085_sp_AD_static_dynamic_def, 028_sp_ranshous2015anomaly_detection_dynamic}. In contrast to static graphs, which only represent a single snapshot of the whole graph $G$, it is easy to detect unusual nodes or vertices using techniques like degree centrality and egonet density \cite{s8_jiang2016_CATCHSYNC}, density-based techniques \cite{s2_you2021_identity_aware}, and others. Dynamic graphs, on the other hand, allow the inclusion of temporal dynamics in the evaluation of vertex behavior \cite{028_sp_ranshous2015anomaly_detection_dynamic}. A formal definition is provided below.

\begin{definition} \textbf{Node Anomaly in dynamic graph} (from \cite{028_sp_ranshous2015anomaly_detection_dynamic}) Given a dynamic graph $G_t$, the total vertex set $V = \cup_{t=1}^T V_t$, and a specified scoring function $f : V \rightarrow \mathbf{R}$, the set of anomalous vertices $V^{\prime} \subseteq V$ is a vertex set such that $\forall v^{\prime} \in V^{\prime}$, $|f(v^{\prime}) - \hat{f}| > c_{0}$, where $\hat{f}$ is a summary statistic of the score $f(v), \forall v\in V$.

\end{definition}

\textbf{Alternative Definition:} Given a dynamic graph $\mathcal{G}_t$ at time $t$, let $t$ belong to a set of discrete time points $T$, and let $v_i$ denote a node in the dynamic graph $\mathcal{G}_t$ at time $t$. We utilize an anomalous scoring function $\Phi_\omega = A(v_i, t)$ to measure the deviation of node $v_i$ at time $t$. Node $v_i$ is considered an \textit{anomalous node} at time $t$ if $A(v_i, t) \geq \theta$. It's important to note that nodes with a scoring function $\Phi_\omega \geq \theta$ are classified as anomalous, and $\theta$ represents the threshold for this classification.

\subsubsection{Edge-level Anomalies:}
Unlike node anomaly detection, edge-level anomaly detection focuses on identifying unusual or irregular patterns in the edges or relationships between elements in a network. A formal definition is provided below.

\begin{definition}\textbf{Edge Anomaly in dynamic graph} (from \cite{028_sp_ranshous2015anomaly_detection_dynamic}) Given a dynamic graph $G_t$, the total edge set $E = \cup_{t=1}^T E_t$, and a specified scoring function $f: E \rightarrow \mathbb{R}$, the set of anomalous edges $E^\prime \subseteq E$ is an edge set such that $\forall e^\prime \in E^\prime, | f(e^\prime) - \hat{f} | > c_0$, where $\hat{f}$ is a summary statistic of the scores $f(e), \forall e\in E$. 
\end{definition}

\textbf{Alternative Definition:} Given a dynamic graph $\mathcal{G}_t = (V, E_t)$, where $V$ is the set of vertices, and $E_t$ is the set of edges at time $t$, each edge $e_{ij}(t) \in E_t$ represents a tuple $(i,j)$ at time $t$. Let the edge-level scoring function be $\Phi_{e_{ij}(t)} = f(e_{ij}(t))$, an edge-level anomaly is detected in $\mathcal{G}_t$ when the scoring function $\Phi_{e_{ij}(t)} > \theta$ exceeds a predefined threshold $\theta$.

\subsubsection{Subgraph-level Anomalies}

In subgraph-level anomaly detection, the focus is on identifying anomalous subgraphs or community structures within the dynamic graph. These structures might exhibit characteristics, behaviors, or patterns that deviate from what is typical in the evolving graph. Some common types of network subgraphs or cliques include triangles, quadrilaterals, and bipartite subgraphs \cite{042_yuan2023_motif}. Subgraph anomalies can be found in a wide range of biological \cite{042_yuan2023_motif}, fruadulent \cite{022_zhang2022e-commerce_fraud_detection_graph}, social \cite{046_eswaran2018spotlight},  technological networks \cite{040b_bhatia2021_MSTREAM}, etc.

\begin{definition}\textbf{Subgraph Anomaly in dynamic graph} (from \cite{028_sp_ranshous2015anomaly_detection_dynamic}) Given a dynamic graph $G_t$, a subgraph set   $H = \cup_{t=1}^T H_t$, where $H_t \subseteq G_t$ and a specified scoring function $f: H \rightarrow \mathbb{R}$, the set of anomalous subgraphs $H^\prime \subseteq H$ is a subgraph set such that $\forall h^\prime \in  H^\prime, | f(h^\prime) - \hat{f} | > c_0$, where $\hat{f}$ is a summary statistic of the scores $f(h), \forall h\in H$. 
\end{definition}

\subsection{New and Emerging Anomaly Variants in Dynamic Graphs}
Recent real-world dynamic social networks have experienced new and emerging anomaly variants beyond the basic types (node, edge, and subgraph anomalies). Here we will explore some additional types of anomalous patterns in dynamic graphs.

\subsubsection{Attribute-based Anomalies:} This analyzes node or edge attributes or features beyond just their connections. Deviations from expected attribute values, such as a sudden change in a user's location or purchase behavior, could indicate anomalies.

In real-world scenarios, diverse sets of information can be modeled as attributed graphs \cite{018_liu2022_BOND_static_attributed_Graph}, incorporating structural relationships and attribute information. Consider the Twitter (or X) social network. For instance, users are connected through various social relationships, and they possess multiple profile details like age, gender, location, and income. This is illustrated in the work of Xuexiong et al.\cite{18r_luo2022_Comga}, titled ``ComGA: Community-Aware Attributed Graph Anomaly Detection." The authors consider graph anomalies in attributed graphs as local, global, and structural anomalies, which makes it beneficial to spot existing structural and complex anomalous nodes. Anomalies in this kind of graph network are different from normal node-level anomalies in both structural and attribute aspects. Hence, we classify this kind of diversity in graph networks as attribute-based anomalies.

\begin{definition}\textbf{Attribute-based Anomaly in Dynamic Graph} 
Given an attributed graph as $G = (V, E, X)$, with the vertex set $V = \cup_{t=1}^T V_t$ and the total edge set $E = \cup_{t=1}^T E_t$. Let $A$ be the set of attributes associated with the vertices $V$, and $f : V \times A \rightarrow \mathbf{R}$ be the specified scoring function that evaluates the vertex-attribute pair. The set of attribute-based anomalous vertices $V^{\prime}_{\text{attribute}} \subseteq V$ is defined such that for every vertex $v^{\prime} \in V^{\prime}_{\text{attribute}}$ and associated attribute $a \in A$, the anomaly score function with a threshold $C_{\text{attribute}}$ is defined as: $|f(v^{\prime}, a) - \hat{f}| > C_{\text{attribute}}$, where $\hat{f}$ is a summary statistic of the score $f(v, a)$ for all $v \in V$ and $a \in A$.

\end{definition}

\subsubsection{Context-aware Anomaly:} This kind of anomaly incorporates additional or contextual information, thereby providing a broader understanding of the dynamic behavior of graph network data, such as time-series data or external events. Context-aware anomalies can manifest in various forms, including temporal information (such as time-series data, timestamps, and seasonal changes) and external events (such as weather conditions, holidays, and news events). Additionally, these anomalies may be observed in multimodal data, such as sensor readings and video footage, as shown in the work of Kim et al. \cite{19r_kim2023_contextual} on contextual anomaly detection for high-dimensional data with a variational autoencoder.

Real-world scenarios of context-aware anomalies are evident in historical and temporal changes in network traffic data \cite{043_F-FADE_chang2021f}, involving factors like packet size and source and destination IP addresses within different timestamps (hours, days, weeks, etc.). Another context-aware anomaly example can be found in user purchase history and product ratings within Recommender Systems \cite{022_zhang2022e-commerce_fraud_detection_graph}. The relevant context in this scenario includes temporal changes and user demographics, such as age, location, and browsing history.

\subsubsection{Community Anomaly:} This kind of anomaly in dynamic graph networks focuses on groups of nodes that exhibit unusual and collective behaviors deviating from the typical community structure of the network. While individual nodes or edges within the community might not be inherently anomalous, the combined deviation of the entire community suggests an anomaly. Real-world scenarios include a sub-community within an online social platform that exhibits abnormal posting patterns \cite{084e_fan2019_graph_social_analysis} and a group of individuals in a specific geographic location showing an unusual pattern of disease infection compared to the surrounding areas \cite{042_yuan2023_motif}.

Community anomalies differ from subgraph anomalies; the latter tends to focus more on specific substructures or subgraphs of the network that deviate from the expected pattern. In contrast, community anomalies focus on the collective behavior of nodes within the network community, considering group-level dynamics.

Another example of community anomalies can be observed in a research group within a collaboration network, which shows significantly fewer connections outside their group domain compared to other research groups, suggesting a lack of collaboration with other groups.

\subsubsection{Multi-Layer Anomaly:} Multi-layer anomalies arise when graph networks have multiple layers of information, where each layer represents a different type of data (e.g., social network connections and communication data). Anomalies might arise from inconsistencies or unusual interactions between the layers. Detecting multi-layer anomalies in dynamic graphs has gained increased attention in recent years. Recent techniques include the 2023 works of Xie et al. \cite{039c1_xie2023_multi_view} on multi-view change point detection, and MultiLAD by Huang et al. \cite{039c2_huang2023laplacian_MultiLAD}. MultiLAD leverages the Laplacian approach to detect change point anomalies in multi-view dynamic graphs. Bhatia et al. \cite{040b_bhatia2021_MSTREAM} also proposed MSTREAM in 2021 for multi-aspect stream anomaly detection. 

Identifying multi-layer anomalies can be challenging due to inconsistent attributes in data, unexpected interactions in graph layers, and the evolving nature of the network. However, in order to detect such anomalies, researchers need to go beyond traditional methods by considering the richness and complexity of multi-layered graph data.

\textbf{Summary: }It is important to note that different types of anomalies in dynamic graphs are network-dependent, and they tend to address diverse aspects, including individual node behavior (node anomalies), relationships and connections (edge anomalies), structural patterns (i.e., an overall change of the graph structure), attribute information (i.e., attribute-based anomalies), community patterns, context of the network, multi-layer interactions, and overall graph structure. Detecting these anomalies requires specialized approaches tailored to the unique characteristics of each anomaly type, emphasizing the need for comprehensive anomaly detection methods in dynamic graph analysis.


\section{Graph Neural Networks Overview}
\label{sec3:GNN_intro}

In this section, we will introduce the concept of a graph neural network (GNN), which is a general architectural framework for modern deep graph learning representations. (\textit{\textbf{Readers already familiar with the architecture of a GNN can skip this section.}}) In 1997, Sperduti et al \cite{g1_sperduti1997_GNN} applied the concept of neural networks to directed acyclic graphs, which sparked the first research in this area.
A GNN was first proposed in the work by Gori et al. \cite{g2_gori2005new_14} in 2005 and subsequently expanded upon in the research by Scarselli and their team in 2009 \cite{g3_scarselli2008graph}, as well as by Gallicchio et al. in 2010 \cite{g4_gallicchio2010graph}. These initial variants are classified as recurrent graph neural networks (RecGNNs).

\subsection{The Basic GNN Architecture}

The core idea of graph neural networks (GNNs) involves generating suitable node representations that depend on the graph structure and feature information. GNNs learn a node's embedding by iteratively encoding its neighboring information into target nodes until a stable fixed vector point is established. The embedding space can be used for several downstream tasks, such as node classification, link prediction, and anomaly detection.

GNNs are designed to work with graph data structures, unlike classical deep learning models such as CNN and LSTM, \cite{083_li_jure2022_power_of_GNN} which are optimized for sequences of images, grids, and text. The basic GNN models have been derived as a generalization of convolutions to non-Euclidean data \cite{g5_hamilton2020graph_representation_textbook}, and they rely on two fundamental principles: the message-passing mechanism and the information-aggregation function.

\begin{figure*}[h]
  \centering
  {\includegraphics[width=0.7\linewidth]{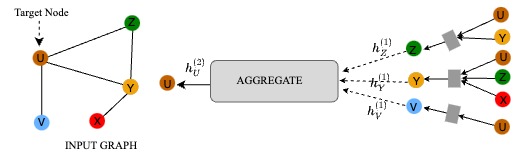}}
  
  \caption{ An Overview of how a single target node $U$ aggregates messages from its local neighborhood (recreated from  Leskovec et al. lecture slide  \cite{fig3_message_parsing}). Given an input graph, the model computes the neighborhood embedding $h_{U}^{(2)}$ by aggregating the messages from $U$'s local neighbors (nodes $Z, Y, V$), and these incoming messages are based on the information aggregated from their respective layers' representations, $h_{Z}^{(1)}$ for node $Z$, $h_{Y}^{(1)}$ for node $Y$, and $h_{V}^{(1)}$ for node $V$. }  
  \label{fig3:GNN_Aggregation}
\end{figure*}

\subsection{GNN Message Passing}
\label{subsec:3.2_message_passing}

In GNNs, message passing is the fundamental mechanism by which information is propagated and aggregated throughout the graph structure to learn representations for nodes or edges. The core concept of message-passing is that, in each iteration, nodes aggregate information from their nearby neighbors \cite{g5_hamilton2020graph_representation_textbook}. As these iterations continue, the node embeddings become increasingly updated about distant portions of the graph, which is often referred to as the "k-hops neighborhood." The "k" in "k-hops neighborhood" refers to the number of hops or steps away from a given node in a graph. The k-hop neighborhood of a node includes the node itself, all its immediate neighbors (1-hop), their neighbors (2-hops), and up to k-hops. In Figure \ref{fig3:GNN_Aggregation}, for node $\textbf{U}$, $k=2$-hops with local neighbors (nodes $Z, Y, V$). In simpler terms, over time, a node's embedding contains information not only about its immediate neighbors but also about the features of nodes further away in the graph.

In every phase of GNNs message passing, a hidden embedding layer ${h}_u^{(k)}$ for each node $u \in V$ undergoes an update that relies on the information accumulated from the neighborhood $\mathcal{N}(u)$ of $u$. A single-layer message passing is depicted in Figure ~\ref{fig3:GNN_Aggregation}, and its mathematical representation is given by Equations \ref{aggregate} and \ref{update}. At the initial phase  $k=0$, the node embedding is ${h}_u^{0} = x_u,  \forall u \in V,$ 

\begin{equation}
m_{\mathcal{N}(u)}^{(k)} = \boldsymbol{f}_{\textit{aggregate}}^{(k)}\left( {h}_v^{(k)} : \forall v \in \mathcal{N}(u) \right),
\label{aggregate}
\end{equation}

\begin{equation}
{h}_u^{(k)} = \boldsymbol{f}_{\textit{update}}\left( {h}_u^{(k-1)}, m_{\mathcal{N}(u)}^{(k)} \right),
\label{update}
\end{equation}
where $m_{\mathcal{N}(u)}^{(k)}$ is the message-passing function that aggregates the neighborhood $\mathcal{N}(u)$ of node $u$, ${h}_u^{(k)}$ is the updated hidden embedding  of node  $u$ at layer $k$, and  ${h}_u^{(k-1)}$ is the hidden embedding of node 
$u $ from the previous layer $k-1$. 
At each step $k$, the aggregation function $\boldsymbol{f}_{\textit{aggregate}}$ in Equation \ref{aggregate} takes the set of embeddings ${h}_v^{(k)}$ for all neighbors $v \in \mathcal{N}(u)$ as input and generates an aggregated message for the neighborhood $\mathcal{N}(u)$.
The update function $\boldsymbol{f}_{\textit{update}}$ in Equation \ref{update} updates the message $m_{\mathcal{N}(u)}^{(k)}$ with the previous embedding $h_u^{(k-1)}$ of node $u$ to generate the current embedding $h_u^{(k)}$.

\subsection{GNN Aggregation}

The aggregation function is also a critical component of the message-passing process as shown in equation \ref{aggregate}. It defines how the node neighborhood information is combined into a single fixed vector or target node.
There are several aggregation methods depending on the GNN variants. The following are a few of them.

\subsubsection{Sum and Mean Aggregation}
These techniques were employed within GraphSAGE \cite{089_hamilton2017_GraphSAGE} GNN variation. In the case of sum aggregation, the information passed between nodes is straightforwardly added up. In mean aggregation, it involves computing the average of messages originating from node \textit{u}'s neighbors. Sum and mean aggregation are simple to implement, computationally efficient, and suitable for capturing global graph properties, as illustrated in the works of Shiyi et al. \cite{31r_huang2023global_GNN} on capturing molecular-level (global) message passing in a GNN. However, sum and mean aggregation may be limited in preserving node-specific information and sensitive to outliers in dynamic graph networks. The mathematical expressions for the sum and mean aggregation are illustrated in Equations \ref{sum_aggr} and \ref{mean_aggr} respectively:

\begin{equation}
    \boldsymbol{f}_{\text{aggregate}}\left( \{ h_u : u \in \mathcal{N}(v) \} \right) = \sum_{u \in \mathcal{N}(v)} h_u,
\label{sum_aggr}
\end{equation}

\begin{equation}
       \boldsymbol{f}_{\textit{aggregate}} \left( \{ h_u : u \in \mathcal{N}(v) \} \right) = \frac{1}{\mathcal{N}(v)} \sum_{u \in \mathcal{N}(v)}h_u,
       \label{mean_aggr}
\end{equation}
where $\boldsymbol{f}_{\text{aggregate}}$ is the aggregation function, and the expression $\{ h_u : u \in \mathcal{N}(v) \}$ is the set of embeddings of the neighbors of node $v$. Here, $h_u$ represents the embedding of neighbor $u$, and $\mathcal{N}(v)$ denotes the set of all neighbors of node $v$. In Equation \ref{sum_aggr}, $\sum_{u \in \mathcal{N}(v)} h_u$ denotes the summation operation over all neighbors $u$ in the neighborhood $\mathcal{N}(v)$ of node $v$, and in Equation \ref{mean_aggr}, $\frac{1}{|\mathcal{N}(v)|}$ is the normalization factor, which is the inverse of the number of neighbors of node $v$. This factor ensures that the mean aggregated embedding is the average of the neighbors' embeddings.

\subsubsection{Graph Convolutional Network (GCN) Aggregation}
This technique was first introduced in the GCN paper by Kipf et al.  \cite{066_GCN_kipf2016semi_supervised_GCN}. GCN aggregations are most suitable for GNN tasks involving the normalization of node-level graph tasks; they have also been shown to be effective in semi-supervised learning tasks \cite{066_GCN_kipf2016semi_supervised_GCN}, as illustrated in Graph Convolutional Extreme Learning Machines by Zhang et al. \cite{32r_zhang2023_semi_GCN}. The major limitation of GCN aggregation is that it struggles with capturing long-range dependencies. The mathematical expression for GCN aggregation is illustrated in Equation \ref{gcn_Aggr} below:

\begin{equation}
    \boldsymbol{f}_{aggregate}\left( \{ h_u : u \in \mathcal{N}(v) \} \right) = \frac{1}{\sqrt{|\mathcal{N}(v)| \cdot |\mathcal{N}(u)|}} \sum_{u \in \mathcal{N}(v)} W h_u,
    \label{gcn_Aggr}
\end{equation}
where $\boldsymbol{f}_{\text{aggregate}}$ is the aggregation function,
the expression $\{ h_u : u \in \mathcal{N}(v) \}$is the set of embeddings of the neighbors of node $v$, $h_u$ denotes the embedding of neighbor $u$, and $\mathcal{N}(v)$ denotes the set of all neighbors of node $v$.
The fraction $\frac{1}{\sqrt{|\mathcal{N}(v)| \cdot |\mathcal{N}(u)|}}$ is the normalization factor, which is the inverse of the square root of the product of the degrees of nodes $v$ and $u$. This factor helps to scale the contributions of neighboring nodes appropriately.
$\sum_{u \in \mathcal{N}(v)} W h_u$ is the summation operation over all neighbors $u$ in the neighborhood $\mathcal{N}(v)$ of node $v$, where $W$ is the learnable weight matrix applied to the embeddings $h_u$ of the neighbors.

The GCN aggregator in Equation \ref{gcn_Aggr} is different from the sum and mean aggregators in Equations \ref{sum_aggr} and \ref{mean_aggr} because it has a normalization factor and a weight matrix that can be learned. This allows it to scale the contributions of neighboring nodes and learn weighted representations, thereby enhancing its ability to capture complex node relationships.

\subsubsection{Graph Attention (GAT) Aggregation}
This approach was initially introduced in the Graph Attention paper by Velickovic et al. \cite{064_0velickovic2017_GAT}. In this method, messages are assigned weights based on attention scores before they are aggregated. GAT aggregations are suitable for tasks where capturing influential node degrees is important, and situations where an attention mechanism is used to improve GNNs, such as node classification with graph attention \cite{064_0velickovic2017_GAT}, node prediction with graph transformers \cite{084f_chen2021_graph_transformer_IoT}, and node and subgraph detection in hybrid-order graphs \cite{060_huang2021_HO-GAT}. Another strength of GAT aggregation is the adaptability in capturing different importance levels of node neighbors. However, they are computationally more expensive compared to simple aggregations (sum and mean), and they could be sensitive to hyperparameter choices. The mathematical expression for GAT aggregation is illustrated in Equation \ref{gat_aggr} below:

\begin{equation}
    f_{\textit{aggregate}}\left( \{ h_u : u \in \mathcal{N}(v) \} \right) = \sum_{u \in \mathcal{N}(v)} \text{softmax}(e_{uv}) h_u,
    \label{gat_aggr}
\end{equation}
where $\boldsymbol{f}{\textit{aggregate}}$ is the aggregation function, ${ h_u : u \in \mathcal{N}(v) }$ represents the set of embeddings of the neighbors of node $v$. Here, $h_u$ denotes the embedding of neighbor $u$, and $\mathcal{N}(v)$ denotes the set of all neighbors of node $v$. The $\text{softmax}(e{uv})$ is the attention coefficient for the edge between nodes $u$ and $v$, computed using the \textit{softmax function}. This coefficient determines the importance of node $u$'s contribution to node $v$. The sum $\sum_{u \in \mathcal{N}(v)}$ represents the summation operation over all neighbors $u$ in the neighborhood $\mathcal{N}(v)$ of node $v$, and $e_{uv}$ is the attention score for the edge between nodes $u$ and $v$, which is typically computed as a function of the embeddings of nodes $u$ and $v$.

The GAT aggregation in Equation \ref{gat_aggr} is superior to mean, sum, and GCN aggregation because it assigns different importance to each neighbor using attention coefficients \cite{10_luong2015_attention_paper}, allowing it to dynamically focus on the most relevant neighbors and capture more nuanced relationships within the graph.

Other aggregation functions, such as LSTM Aggregation employed for sequential message passing, were utilized in GraphSAGE \cite{089_hamilton2017_GraphSAGE}. The LSTM aggregation is mostly suitable for dynamic graph tasks with temporal dependencies and is also applicable in sequential tasks in graph representation learning. This is illustrated in AddGraph by Zheng et al. \cite{026_AddGraph_zheng2019addgraph} for capturing node and edge structural information and temporal dependencies in evolving graphs. However, they could be computationally intensive compared to simple aggregations.

It is important to emphasize that the choice of aggregation function is dependent on the GNN framework utilized and the nature of the graph-related problem at hand. In the literature, researchers often explore various functions to enhance performance.

Basic GNN Message Passing: Equations \ref{aggregate} and \ref{update} offer a high-level perspective on the Aggregation and Update functions within GNN frameworks. The message passing mechanism in the original GNN model, as introduced by Gori et al. \cite{g2_gori2005new_14} and Scarselli et al. \cite{g3_scarselli2008graph}, is formally expressed in Equation \ref{gnn_framework}:

\begin{equation}
\label{gnn_framework}
h_u^{(k)} = \sigma\left( {W}_{\textit{self}}^{(k)} h_u^{(k-1)} + {W}_{\textit{neighbor}}^{(k)} \sum_{v \in \mathcal{N}(u)} h_v^{(k-1)} + {b}^{(k)} \right),
\end{equation}
where $h_u^{(k)}$ denotes the updated embedding of node $u$ at layer $k$, $h_u^{(k-1)}$ is the node embedding of $u$ from the previous layer $k-1$, ${W}_{\textit{self}}^{(k)}$ and ${W}_{\textit{neighbor}}^{(k)}$ are trainable weights in $\mathbb{R}^{d^{(k)} \times d^{(k -1)}}$, and $\sigma$ represents the non-linear function, such as ReLU or tanh. As in other deep learning models, $b^{(k)}$ serves as the bias term at layer $k$.
The key idea underlying basic GNN message passing, as described in Equation \ref{gnn_framework}, is its analogy to the standard multi-layer perceptron (MLP) \cite{g5_hamilton2020graph_representation_textbook}. This analogy stems from its reliance on linear operations followed by a single element-wise non-linearity.

For a deeper understanding of the GNN framework, including the intricacies of message passing, self-loop operations, and generalized neighborhood aggregation in Graph Neural Networks (GNNs), we recommend referring to the work by Hamilton \cite{3c_sp_kazemi2020representation_dynamic_graph} which provides comprehensive insights into these concepts.

Deep Graph Neural Network (GNN) node embedding representations have demonstrated remarkable success in tackling various network-related tasks. Some of these tasks are node classification, which involves labeling nodes with their corresponding categories; link prediction, which detects patterns in densely connected node clusters; and network similarity assessment, which measures how similar are different sub-networks.



\section{Dynamic Graph Representation}
\label{sec4:dynamic_graph_rep}
Dynamic graph networks in real-world scenarios include social networks (facilitating the spread of news or information among friends), transportation (monitoring traffic flow on roads), financial (tracking the movement of money through an economy), network traffic, electricity grid dynamics, and biological processes. These networks can be represented in diverse ways, and the success of graph learning tasks relies heavily on the topology or structure of the networks, specifically the arrangement of nodes and edges \cite{071_sp_skarding2021_foundations_dynamic_graph_survey}.

Dynamic systems come in discrete-time and continuous-time forms and may exhibit either deterministic or stochastic characteristics \cite{03_tb_newman2018_Networks_mark_newman}. In these sections, we provide details on the dynamic graph representation.
    

\begin{table*}[htbp]
  \caption{Comparison of Dynamic Graph Representations: Discrete vs. Continuous Networks}
  \label{tab:dynamic_rep_comparison}
  \begin{tabular}{p{1.5cm} p{5cm} p{4.5cm}}
    \toprule
    \textbf{Dynamic layouts} & \textbf{Temporal Properties} & \textbf{Network Types}\\
    \midrule
    Discrete & Distinct time \& Fixed intervals, Sparse changes, represent abstract relationships & Snapshots, Time Slices, incremental updates  \\
    \midrule
    Continuous & Event-based, Continuous evolution, represent spatial relationships & Graph Streams, Transitioning graphs \\
    \bottomrule
  \end{tabular}
\end{table*}

\subsection{Discrete Representation}
\label{subsec:discrete}

A discrete graph representation tends to model a system where the relationships between entities change over time in a discrete manner \cite{03_tb_newman2018_Networks_mark_newman}. In such graphs, the structure of the graph evolves at distinct time steps, or snapshots, to capture the dynamic nature of the network. A discrete representation is illustrated in Figure \ref{fig1:dynamic_representation} and shown in Equation 4 as 

\begin{equation}
    \mathcal{G} = \{ G^1, G^2, \dots, G^T \},
\end{equation}
where $G^i$ represents graph snapshots, and ${T}$ denotes the sequence of time steps for each snapshots.

Modeling dynamic networks as graph snapshots allows for static analysis at individual time steps and, collectively, provides insights into the entire network \cite{071_sp_skarding2021_foundations_dynamic_graph_survey}. Several dynamic graph algorithms capture snapshots using techniques such as sliding windows \cite{042_yuan2023_motif}, multi-layered networks \cite{083_you2022_Roland,115_goyal2018_DynGem,114_liu2021_TADDY}, the spectrum of Laplacian matrix or tensors \cite{039_huang2020laplacian_LAD}, first-order Markov process \cite{039b_wang2017_Edge-Monitoring}, and many more. See Section \ref{sec5:Ad_approaches} for details on how these techniques apply the snapshot approach in modeling dynamic graphs.

\subsection{Continuous Representation}
\label{subsec:continuous}

A continuous graph representation, on the other hand, extends traditional graph structure to model systems where relationships between entities evolve continuously over time as opposed to discrete time steps. A continuous representation captures exact temporal information and is more complex to model mathematically \cite{071_sp_skarding2021_foundations_dynamic_graph_survey}. This representation is particularly relevant in applications such as neuroscience, physics, the spread of infectious diseases, and social dynamics, where changes in connections or attributes of graph elements occur smoothly and continuously. We illustrate the evolution of a continuous graph in Figure \ref{fig1:dynamic_representation}. 

Mathematically, the continuous evolution of graph networks is frequently modeled based on network topologies, and the continuous graph evolution can be described through differential equations, integrals, or other mathematical frameworks.
Let $\mathcal{G}(t)$ be a continuous graph representation with a set of nodes $V(t)$ and edges $E(t)$ at time $t$. The evolution of a graph can be modeled by a system of differential equations

\begin{equation}
    \dv{V}{t} = f_V (V, E, t),  
\end{equation}

\begin{equation}
    \dv{E}{t} = f_E (V, E, t),  
\end{equation}
\begin{equation}
    \dv{W}{t} = f_W (V, E,W, t),  
\end{equation}
where $W(t)$ represents the edge weight function at time $t$, $f_V$, $f_E$, and $f_W$ are functions describing the continuous changes of vertices, edges, and edge weights, respectively. For more details, we have highlighted the comparison of discrete and continuous dynamic graph representations in Table~\ref{tab:dynamic_rep_comparison}.

\subsection{Hybrid Representation}
In subsections \ref{subsec:discrete} and \ref{subsec:continuous}, we introduced two distinct approaches to dynamic graph representation: discrete and continuous. While discrete representation tends to model the qualitative aspects of the network interactions and transitions per timestep, continuous representation focuses on modeling the quantitative aspects of the dynamic graph entities as they evolve continuously over time. This includes capturing evolving and temporal information over timestamps as opposed to discrete time snapshots.

In emerging complex graph networks, there is a possibility of a \textbf{hybrid representation} that serves as a bridge between the qualitative and quantitative dimensions of graph networks. A hybrid graph representation allows for the seamless integration of both discrete and continuous information. This mapping facilitates a holistic understanding of dynamic graph behavior, enhancing the ability to detect anomalies that may manifest in various forms.

Real-world instances of hybrid representations include (1) Event-driven anomaly detection in financial systems by considering discrete irregularities and continuous fluctuations in market behavior; this is illustrated in the works of Wu et al. \cite{27r_wu2021event2_Event2graph_timeWu} on multivariate time-series. (2) Anomaly detection and fault diagnosis in smart grids by considering both discrete disruptions and continuous variations in power grid systems; this was shown in the works of Li et al. \cite{030_li2021_DYNWATCH_power_grid} DYNWATCH in 2021. (3) Sensor network detection in environmental monitoring by considering both sudden changes in sensor status and temporal variations \cite{032_sp_kyle2023survey_AD_IoT_Sensor_Networks}. This scenario can also be found in cyber-physical systems for detecting polymorphic malware (malicious attacks that can change its code) and intrusion attempts; this is shown in the works of Jeffrey et al. \cite{26r_jeffrey2024_hybrid_cyber_physical} in 2024.

\section{Anomaly Detection Methods}
\label{sec5:Ad_approaches}

In this section, we aim to introduce and compare different \textbf{D}ynamic \textbf{G}raph-Based \textbf{A}nomaly \textbf{D}etection (\textbf{DGAD}) methods. 

In our survey approach, we categorize our DGAD methods into four major groups based on their respective approaches and core algorithms for identifying anomalous patterns in dynamic graphs. These groups include traditional machine learning-based, matrix transformation, probabilistic, and deep learning methods. We further subdivide these categories to provide a more narrow description of the algorithms. It is important to note that certain methods in our survey topology in Figure ~\ref{fig2:survey_framework} and Table \ref{tab2_AD_approach} may overlap with other groups. Nevertheless, our survey aims to capture recent trends in anomaly detection techniques for dynamic graphs while also highlighting commonly used dynamic graph datasets and evaluation metrics.

The summary of the current papers, paper descriptions, specific graph learning tasks, the datasets utilized, and the evaluation metrics are presented in Table  ~\ref{tab2_AD_approach}.

\subsection{Traditional Machine Learning Methods}
Traditional machine learning (ML) methods for anomaly detection involve the use of established algorithms and techniques \cite{024_sp_ma2021comprehensive_survey_AD_Deeplearning}. These methods rely on predefined rules or patterns to identify anomalies in datasets. Examples of these techniques include statistical, tree-based, clustering, distance-based approaches, and many others. Over the decade, traditional ML techniques have proven to be effective for many downstream tasks, such as anomaly detection and link prediction in graphs; however, they are faced with challenges in handling high-dimensional or complex graph data, and more advanced methods are often considered in such cases. 

In our survey, we categorize the traditional ML methods into tree-based, density-based, and distance-based. (See Table \ref{tab2_AD_approach} for details.)

\subsubsection{Tree-based and Density-based Methods}
Tree-based and density-based anomaly detection methods are two distinct approaches for detecting anomalies. Tree-based methods involve constructing a decision tree or an ensemble of decision trees, such as Random Forest or Isolation Forest, to isolate and identify anomalies \cite{AD2_na2018_DILOF, AD3_salehi2016_fast_LOF}. While density-based methods focus on identifying anomalies based on the density of data points in the feature space. Density-based models include DBSCAN (Density-Based Spatial Clustering of Applications with Noise), LOF (Local Outlier Factor), and One-Class SVM (Support Vector Machine) \cite{AD4_liu2008_isolation_Forest, AD5_hariri2019_extended_Isolated_Forest}.

The Local Outlier Factor (LOF) \cite{AD1_2000_LOF} is one of the most popular density-based algorithms. It works by measuring the local density of each data point and identifying anomalous points with significantly lower densities compared to their neighbors. This approach is mostly used in scenarios where the traditional distance-based approach may not perform well, such as non-uniformly distributed data points. MiLOF \cite{AD3_salehi2016_fast_LOF}, an incremental local outlier detection algorithm, expands LOF for data streams.
To address the memory issue and the limitation of detecting long sequences of outliers, DILOF \cite{AD2_na2018_DILOF} improved upon the LOF \cite{AD1_2000_LOF} and MiLOF \cite{AD3_salehi2016_fast_LOF} algorithms by adopting a novel density-based sampling algorithm to summarize past data and a new strategy for detecting outlier sequences. Recently, Goodge et al. \cite{120_goodge2022_LUNAR} introduced LUNAR, a hybrid approach that combines deep graph neural networks (GNN) and LOF to learn information from the nearest neighbors of each node in a trainable manner for anomaly detection. 
However, these techniques are most effective when dealing with data of lower dimensions, as they are susceptible to the curse of dimensionality when applied to higher-dimensional data.
\subsubsection{Distance-based}
Distance-based anomaly detection techniques in dynamic graphs propose certain time-evolving measures of dynamic network structures and leverage the change rates of those measures to detect anomalies. These methods focus on tracking how network properties evolve over time and identifying deviations indicative of unusual network behavior.

StreamSpot \cite{038_manzoor2016_fast_StreamSpot} is a clustering-based AD approach that utilizes a novel similarity function for heterogeneous graphs in real-time from a continuous stream of typed edges. This framework is tailored to process temporal graphs with categorically designated nodes and edges while simultaneously upholding the efficacy of graph sketches and clustering configurations. StreamSpot employs a shingling-based similarity function to create graph sketches that capture structural information, enabling memory-efficient comparisons. In addition, StreamSpot further encompasses strategies for the progressive upkeep of these sketches and clustering arrangements, adapting to the dynamic nature of incoming edge data. Empirical validation of StreamSpot is conducted via quantitative assessments on synthesized datasets encompassing both normal and abnormal activities. StreamSpot obtains over a $90\%$ average precision on approximately $25M$ system log streaming edges, while the overall performance decreases as memory is constrained (i.e., detection is delayed). However, the running time and recovery decays are slow for high-volume streams, making the approach less scalable.

Eswaran Dhivya et al. \cite{101_eswaran2018_SedanSpot} proposed SedanSpot, a randomized algorithm for anomaly detection in edge streams. It uses a holistic random walk-based edge anomaly scoring function to compare an incoming edge with the whole (sampled) graph, emphasizing the importance of far-away neighbors. SedanSpot detects edges that connect sparsely connected parts of a graph, and it identifies edge anomalies based on edge occurrence, preferential attachment, and mutual neighbors. As an improvement to SedanSpot, Eswaran et al. \cite{046_eswaran2018spotlight} proposed SpotLight, a randomized sketching-based method for detecting sudden changes in dynamic graphs and detecting the appearance and disappearance of dense subgraphs or bi-cliques using sketching. SpotLight guarantees that an anomalous graph is mapped ‘far’ away from ‘normal’ instances in the sketch space with a high probability for an appropriate choice of parameters. This is done by creating a K-dimensional sketch that comprises K subgraphs, thereby enabling the detection of sudden changes within the dynamic graph.

Compared to SedanSpot \cite{101_eswaran2018_SedanSpot}, which relies on a random walk algorithm, SpotLight \cite{046_eswaran2018spotlight} uses a randomized sketch algorithm, making it more scalable, fast, and reliable for the identification of the sudden appearance of anomalies in densely directed subgraphs. Experimenting on $1207$ graph snapshots and $288$ ground truth anomalies ($28\%$ of total), SpotLight gave precisions of $(0.79, 0.64, 0.57)$ at cut-off rank $(200, 300, 400)$ respectively and an overall AUC of $0.7$. This is an $8\%$ improvement on the state-of-the-art in 2018. SedanSpot, on the other hand, gave an AUC score of $0.63$ when processing $2.54$ million edges in 4 minutes, and SedanSpot's input stream is processed linearly, resulting in high computational challenges.  AnomRank \cite{027_AnomRank_yoon2019fast} introduced two-pronged approaches for capturing both structural and edge weight anomalous changes. However, AnomRank needs to compute a global PageRank, which does not scale for edge stream processing.

Li et al.\cite{030_li2021_DYNWATCH_power_grid} developed DYNWATCH, 
a distance (or similarity) based approach for real-time anomaly detection using sensor data from the electric power grid. The DYNWATCH algorithm is domain-specific and topology-aware, allowing it to adapt to rapid changes in historical graph data. DYNWATCH \cite{030_li2021_DYNWATCH_power_grid} constructs a graph from the active devices of the grid, using active grid buses as vertices and active grid devices as nodes. It calculates the graph distance using the Line Outage Distribution Factors (LODF) sensitivity measure and performs temporal weighting based on the graph's distance and weights for anomaly detection. In essence, the algorithm works by defining graph distances based on domain knowledge and estimating a reliable distribution of measurements at time $t$ from the most relevant previous data. Other recent distance-based methods include SnapSketch \cite{119_paudel2020_SnapSketch}, a sketching approach that uses a simplified hashing of the discriminative shingles generated from a biased-random walk. DynAnom \cite{113_guo2022_DynAnom} tracks anomalies at both the node and graph levels in large, dynamically weighted graphs, and SOM-based \cite{074_lamichhane2022_SOM-based} clusters visualize the normal and abnormal patterns in graph streams using self-organized maps (SOM).

\subsection{Matrix Factorization and Tensor Decomposition Approach}

\textbf{Matrix factorization} is a mathematical technique that decomposes high-dimensional matrices into lower-dimensional forms. It is applied to model evolving relationships in dynamic graphs, revealing patterns and anomalies over time by factorizing the adjacency matrix. \textbf{Tensor decomposition}, on the other hand, extends matrix factorization to multi-dimensional arrays or tensors. This technique finds latent factors and temporal patterns in dynamic graph data. It is then possible to detect anomalies by decomposing the multidimensional tensors that show how nodes interact over time.

Wang et al. \cite{039b_wang2017_Edge-Monitoring} proposed an Edge-Monitoring technique based on the Markov Chain Monte Carlo (MCMC) sampling theory. Wang et al. modeled the dynamic network evolution as a first-order Markov process. They make the assumption that an unknown foundational model exists that dictates how the generation process works. Additionally, both the current generative model and the previously observed snapshot have an impact on each snapshot of the graph. Their approach is regarded as one of the best for change point detection. However, the major limitation of this approach is that the Edge-Monitoring \cite{039b_wang2017_Edge-Monitoring} relies on consistent node orderings across all time steps. In addition, edge monitoring assumes a constant number of nodes for each snapshot. This assumption can be easily violated in the case of large social networks, which frequently witness the addition of user accounts. 

To address the limitation in \cite{039b_wang2017_Edge-Monitoring}, Huang et al. \cite{039_huang2020laplacian_LAD} introduced LAD (Laplacian Anomaly Detection) which computes the singular value decomposition (SVD) of the graph Laplacian to obtain a low-dimensional graph representation. LAD takes snapshots of the graph structure at different time steps and then applies the spectrum of the Laplacian matrix to make embeddings with low dimensions. The core idea of LAD\cite{039_huang2020laplacian_LAD} is to detect high-level graph changes from low-dimensional embeddings (called signature vectors). The normal pattern of the graph is extracted from a stream of signature vectors based on both short-term and long-term dependencies, thereby comparing the deviation of the current signature vector from normal behavior. The method addresses two primary challenges in the identification of change points in dynamic graphs: the evaluation of graph snapshots across time and the representation of temporal dependencies. By using the single values of the Laplacian matrix and adding two context windows, LAD makes it possible to compare the current graph to both short-term and long-term historical patterns.

In contrast to Edge-Monitoring \cite{039b_wang2017_Edge-Monitoring}, LAD \cite{039_huang2020laplacian_LAD} exhibits the ability to manage a fluctuating number of nodes over time in the dynamic graph or network. LAD takes this into account by explicitly modeling both short-term and long-term behaviors within the dynamic graph, effectively aggregating the information from both temporal perspectives. Past studies have also focused on detecting anomalous dense subtensors in tensor data, such as social media and TCP dumps. The works of Faloutsos et al. \cite{105_shin2017_DenseAlert} have made significant contributions to the application of the tensor decomposition approach to dynamic graphs. They proposed Fast Dense-Block \cite{d5_shin2016m_FastDense_Wiki_data} and DenseAlert \cite{105_shin2017_DenseAlert}—an incremental and constantly updating algorithm designed for identifying sudden subtensors that emerge within a short time frame.

Unfortunately, the laplacian matrix approaches are computationally expensive, require manual extraction of the dynamic graph properties, and are also susceptible to noise. Xie et al. \cite{039c1_xie2023_multi_view} recently published MICPD, a multi-view feature interpretable change point detection method based on a vector autoregressive (VAR) model to turn high-dimensional graph data into a low-dimensional representation. MICPD finds change points by following the evolution of multiple objects and how they interact across all time steps. Huang et al \cite{039c2_huang2023laplacian_MultiLAD} recently proposed MitliLAD \cite{039_huang2020laplacian_LAD} as a simple and scalable extension of the LAD algorithm to multi-view graphs that finds change points in multi-view dynamic graphs.

\begin{table*}
  \caption{A Comparison of Anomaly Detection (AD) Methods in Dynamic Graphs: A Review of 53 Recent Papers (2016-2023)}
  \label{tab2_AD_approach}
  
  \begin{adjustbox}{center}
  \resizebox{\textwidth}{!}{
    \begin{tabular}{l l l l l lll}
      \toprule
         \multicolumn{2}{c}{\textbf{Methods}} & \textbf{Paper} & \textbf{Year} & \textbf{Summary and Focus of Paper} & \textbf{Learning Task} & \textbf{Dataset} & \textbf{Metrics} \\

          \midrule

     \multirow{15}{*}{\rotatebox[origin=c]{90}{Traditional Machine Learning Methods}}& \multirow{2}{*}{\shortstack{Tree-based}}  &  RRCF \cite{124_guha2016_RRCF} & 2016 & A Robust Random Cut Forest-based AD algorithm in streams  & Node & NYC, Synthetic& ACC, Prec,AUC  \\ 
        &  & Extended-IF \cite{AD5_hariri2019_extended_Isolated_Forest} & 2021 & {\parbox{9cm}{Extends Isolation Forest(IF) \cite{AD4_liu2008_isolation_Forest} where the split is based on hyperplanes with random slopes instead single variable threshold}} & Node & Single-Blob, Sinusoid&AUC-ROC/PRC  \\
         \cline{3-8} \\
        & \multirow{3}{*}{Density-based} & MiLOF \cite{AD3_salehi2016_fast_LOF}  & 2016 & An incremental LOF detection algorithm for data streams & Node & UCI, IBRL, Synthetic& ROC-AUC  \\ 
        &  & DILOF \cite{AD2_na2018_DILOF} & 2018 & {\parbox{9cm}{Improve on LOF and MiLOF using a new density sampling algorithm to summarize the data.}} & Node & UCI, KDDCup99& AUC \\ 
        &  & LUNAR \cite{120_goodge2022_LUNAR} & 2021 & A hybrid  combination of deep learning and LOF & Node & HRSS, MI-F, Shuttle& AUC  \\
         &  & EvoKG \cite{047_park2022_EvoKG} & 2022 & Models the event time by estimating its conditional density & Edge & ICEWS18, Wiki, Yago& MRR, Hits$@$n  \\
        \cline{3-8} \\
        & \multirow{5}{*}{Distance-based} &  StreamSpot \cite{038_manzoor2016_fast_StreamSpot} & 2016 & Anomaly detection in streaming heterogeneous graphs & Node, Edge & Youtube, Email&  ROC-AUC  \\
        & & SpotLight \cite{046_eswaran2018spotlight} & 2018 & Detects sudden (dis)appearance of densely directed subgraph. & Edge, Subgraph & DARPA, ENRON, NYC& Prec., Rec., AUC \\
        &  & SedanSpot \cite{101_eswaran2018_SedanSpot} & 2018& Detects edges that connect sparsely-connected parts of a graph.
     & Edge & DARPA, DBLP, ENRON& Prec., Rec., AUC  \\
        &  & AnomRank \cite{027_AnomRank_yoon2019fast} & 2019& {\parbox{9cm}{Detecting anomalies in dynamic graphs with a two-pronged approach.}} & Node, Edge & DARPA, ENRON, Syn. & ACC, Prec.,  \\
        &  & SnapSketch \cite{119_paudel2020_SnapSketch} & 2020 & {\parbox{9cm}{Shingling technique and baised random walk to sketch the graph}} & Graph & DARPA, IOT-data &Prec, Rec.\\
        &  & DYNWATCH \cite{030_li2021_DYNWATCH_power_grid} & 2022 & Anomaly detection using sensors placed on a dynamic grid & Edge, Graph &Grid data (private)& ROC-AUC, F-1 \\
        &  & DynAnom \cite{113_guo2022_DynAnom}  & 2022 & Detect anomalies in large, dynamically weighted graphs & Node, Edge, Graph & DARPA, EuCore, ENRON& Precision  \\ 
        &  & SOM-based \cite{074_lamichhane2022_SOM-based}   & 2022 & {\parbox{9cm}{A self-organized map (SOM)-based clustering and visualization approach on streaming graphs}} & Node, Graph & AST2012, UNSW, ISCX& t-SNE Maps  \\ 
        \hline \\

\multirow{5}{*}{\rotatebox[origin=c]{90}{\shortstack{Matrix-TF}}}& \multirow{4}{*}{\shortstack{Matrix/Tensor \\ Decomposition}}  & EdgeMonitor \cite{039b_wang2017_Edge-Monitoring} & 2017 & It models dynamic graph as a first order
Markov process & Edge & Synthetic, Senate& Rec., Prec.  \\

    & & DenseAlert \cite{105_shin2017_DenseAlert} & 2017 & Detecting dense subtensor in tensor stream  & Sub-graph & {\parbox{4cm}{Rating (Yelp, Android), \\ KoWiki,  Youtube, DARPA}}   & Density, Rec. \\
       & & Laplacian-AD \cite{039_huang2020laplacian_LAD} & 2020 & Laplacian spectrum  for change point detection. & Node, Subgraph & Synthetic UCI, Senate & Hits$@ n$ \\
        &  & MICPD \cite{039c1_xie2023_multi_view}& 2023 & Interpretable change point detection  method & Node, Graph & Synthetic, World Trade  & T2 chart \\
        &  & MultiLAD \cite{039c2_huang2023laplacian_MultiLAD} & 2023 & Generalization of LAD \cite{039_huang2020laplacian_LAD} to multi-view graphs & Node, Subgraph & {\shortstack{UCI, Senate, Bill-voting}}  & Hits$@ n$ \\
        \hline \\
\multirow{9}{*}{\rotatebox[origin=c]{90}{\shortstack{Probabilistic methods}}}& \multirow{6}{*}{}  & CM-Sketch \cite{111_ranshous2016scalable_CM-Sketch} & 2016 & Sketch-based outlier detection in edge streams.  & Edge & IMDB, DBLP& AUC  \\
     &  & EdgeCentric \cite{072_shah2016_EdgeCentric}  & 2016 & {\parbox{9cm}{Uses Minimum Description Length to rank node anomalies based on patterns of edge-attribute behavior in an unsupervised way. }} & Edge & {\parbox{3.3cm}{Flipkart, Software Marketplace (SWM)}} & Precision\\

        &  & PENminer \cite{0103_belth2020_PENminer} & 2020 & {\parbox{9cm}{Explores the persistence of activity snippets, i.e., the length and regularity of edge-update sequences’ reoccurrences.}} & Edge & {\parbox{3.6cm}{EuEmail, NYC, DARPA, Boston-Columbus Bike, Reddit, Stackoverflow}}& AUC\\
        
        &  & MIDAS \cite{040_bhatia2022_MIDAS_latest} & 2020 & {\parbox{9cm}{Detects microcluster anomalies in edge streams and uses count-min sketches (CMS) to count edge occurrences.}} & Edge& {\parbox{3.8cm}{ TwitterSec-WorldCup DARPA, CTU13, UNSW15 }} & ROC-AUC \\
        &  & F-FADE \cite{043_F-FADE_chang2021f} & 2021 & {\parbox{9cm}{Frequency-factorization  to detect edge streams anomalies}}  & Edge & {\parbox{3.6cm}{RTM-Synthetic, DARPA, DBLP BARRA, ENRON}}& AUC \\ 
        &  & MSTREAM \cite{040b_bhatia2021_MSTREAM} & 2021 & Detects group anomalies in multi-aspect data & Subgraph& {\parbox{3.7cm}{KDD99, UNSW15, CICIDS}}& ROC-AUC  \\
        &  & AnoEdge \cite{056_bhatia2023_AnoEdge_sketch_based}  & 2023 & {\parbox{9cm}{Detects edge and graphs anomalies by extending the CMS structure in MIDAS \cite{040_bhatia2022_MIDAS_latest} to a Higher-Order Sketch}} & Edge, graph & {\parbox{3.8cm}{DARPA, ISCX-IDS12, CIC-IDS18, CIC-DDoS2019}} & ROC-AUC \\
        
        \hline \\
      \multirow{25}{*}{\rotatebox[origin=c]{90}{Deep Learning Methods}}& \multirow{4}{*}{AutoEncoder}  &  DynGEM \cite{115_goyal2018_DynGem} & 2018 & {\parbox{9cm}{It utilizes deep auto-encoders to incrementally generate embedding of a dynamic graph at each snapshot}} & Edge & HEP-TH, AS, ENRON& Avg. MAP  \\
       & & Dyngraph2vec \cite{116_goyal2020_Dyngraph2vec} & 2020 & Uses multiple non-linear layers to learn structural patterns. & Edge & HEP-TH, AS-dataset&Avg. MAP  \\
        &  & H-VGRAE \cite{057_yang2020_H-VGRAE}  & 2020& uses a hierarchical variational graph recurrent autoencoder & Node, Edge & UCI, HEP-TH, GitHub& AUC  \\
        &  & DGAAD \cite{117_gao2022anomaly_DGAAD} & 2022 & A deep graph autoencoder to learn dynamic node embedding& Node & EuEmail & AUC, ACC, Rec. \\ 
        \Vcline{3-8} \\
        & \multirow{6}{*}{\shortstack{Graph \\ Embedding}} & Node2Vec \cite{grover2016_Node2Vec} & 2016 & {\parbox{9cm}{Uses BFS and DFS in the generation of random walks for learning continuous feature representation.}} & Node, Edge & Facebook, PPI, arXiv& AUC  \\
         & &NetWalk \cite{059_NetWalk_yu2018netwalk} & 2018 & {\parbox{9cm}{Learns network representations for node and edges, and detects  deviations based on a dynamic clustering algorithm.}} & Node, Edge & UCI, Digg, DBLP& AUC  \\
        &  & GraphSAGE \cite{089_hamilton2017_GraphSAGE} & 2018& Inductive representation learning on large graphs & Node & Citation, Reddit, PPI& Micro-avg. F1  \\
        &  & AER-AD \cite{036_fang2023_AER_anonymous_edge}  & 2023 & Inductive anomaly detection in dynamic graphs &Edge & {\parbox{3.5cm}{Mooc Reddit, Amazon, Enron, Wiki}}& F1, AUC  \\
        &  & GraphEmbed \cite{054_wang2023_GraphEmbed}  & 2023 & {\parbox{9cm}{Graph-level embedding that utilized a modified random walk with temporal backtracking}} &Graph & {\parbox{3.5cm}{Reddit, Enron, Facebook, Slashdot}}& Precision \\
         &  & TEST \cite{055_cedre2023_TEST}  & 2023 & Temporal Egonet-subgraph transitions embedding method & Node, Subgraph & Enron, UCI, EuEmail& Prec, Rec, F1  \\

        \Vcline{3-8} \\
        & \multirow{7}{*}{ \shortstack{Deep Graph \\  Learning}  } &  AddGraph \cite{026_AddGraph_zheng2019addgraph} & 2019& Detects edge anomaly with extented GCN, Attention, and GRU & Edge & UCI, Digg & AUC \\ 

        &  & GENI \cite{049_park2019_GENI} & 2019 & GNN-based approach for estimating node importance in KGs  & Node & fb15k, music10k, IMBD   & NDCG \\
        &  & HOLS \cite{048_eswaran2020_HOLS_higher-order-label} & 2020& {\parbox{9cm}{Uses higher-order structures for graph semi-supervised learning}} & Node, Subgraph & {\parbox{3.5cm}{EuEmail, PolBlogs,Cora}}&ACC  \\ 
        
        &  & StrGNN \cite{031_StrGNN_cai2021structural} & 2021 & Leverage structural GNN to detect anomalous edges  & Edge & {\parbox{3.3cm}{UCI, Digg Email-DNC, Bitcoin-Alpha/OTC}}   & AUC \\
        &  & CGC \cite{044_park2022_CGC} & 2022& {\parbox{9cm}{Contrastive learning for deep graph clustering in 
 time-evolving networks}} & {\parbox{2.3cm}{Node, Subgraph, Graph}} & {\parbox{3.6cm}{ACM,DBLP,Citeseer, MAG-CS, NYC, Yahoo}} & {\parbox{1.6cm}{ ACC, NMI, F1, ARI}}  \\
        
        &  & ROLAND \cite{083_you2022_Roland} & 2022 & Extends static GNNs to capture dynamic graphs. & Node, Edge &{\parbox{3.5cm}{Reddit, AS-733, BSI-ZK, UCI, Bitcoin}}& MRR \\
       
       & & PaGE-Link \cite{050_zhang2023_PageLink} & 2023 & GNN explanation for heterogeneous link prediction& Edge & AugCitation & ROC-AUC \\
        & & MADG \cite{042_yuan2023_motif} & 2023 & Motif detection with augmented GCN and self-attention& Subgraph & {\parbox{3.5cm}{UCI,Email-DNC, Bitcoin-Alpha/OTC}}&Prec, Rec.,AUC\\
         &  & SAD \cite{053_tian2023_SAD} & 2023 & {\parbox{9cm}{A semi-supervised AD on dynamic grapp, it uses statistical distribution of unlabeled samples as the reference for loss calculation.}} & Node & Wiki, Reddit, Alipay&AUC\\

        \Vcline{3-8} \\
        & \multirow{8}{*}{\shortstack{Graph \\Attention $\&$\\ Transformer}} &
       GAT \cite{064_0velickovic2017_GAT}& 2018 & An attention-based architecture to perform node classification & Node & Cora, Pubmed, PPI& ACC. F1  \\
         & & HAN \cite{080b_wang2019_HAN}  & 2019 & A heterogeneous GNN based on the hierarchical attention&Node & DBLP, ACM, IMDB & ACC, NMI  \\
         
        & & GTN \cite{080_yun2019graph_Transformer_paper} & 2019 & Graph transformer networks, to learn a new graph structure & Node & DBLP, ACM, IMDB & ACC.  \\
         & & DySAT \cite{061_sankar2018_DySAT} & 2019 & Dynamic graph  learning with self-attention Network & Graph &Enron, UCI, Yelp & AUC  \\
           & & DyHAN \cite{062_yang2020_DyHAN}  & 2020 & {\parbox{9cm}{Dynamic Heterogeneous graph embedding with  Attentions mechanism}} &Node, Edge & EComm, Twitter, Alibaba & ACC,AUC\\
        & & HO-GAT \cite{060_huang2021_HO-GAT} & 2021 & {\parbox{9cm}{A hybrid-order graph attention method for detecting node and subgraph anomaly in a dynamically attributed graph.}} & Node, Subgraph & Scholat, AMiner, WebKB  & Prec, Recall   \\
        & & TADDY \cite{114_liu2021_TADDY} & 2021 & Transformer-based AD model for Dynamic graphs & Node, Edge & {\parbox{3.5cm}{UCI, Alpha, OTC,Digg, EmailDNC, AS-Topology}} &  AUC \\
        &  & Graphormer \cite{063_ying2106_Graphomer_Do_transformers_Bad_Graph} & 2021 & Uses  transformer \cite{17_vaswani2017attention_is_all_you_need} model for graph representation learning & Node, Edge & OGB dataset& MAE, AUC \\

      \bottomrule
      \multicolumn{8}{l}{*\textbf{ACC}: Accuracy, \textbf{NMI}: Normalized Mutual Information, \textbf{ARI}: Adjusted Rand Index, \textbf{ROC-AUC}: Area Under the Receiver Operating Characteristic Curve} \\
      \multicolumn{8}{l}{*\textbf{ROC-AUC}: Area Under the Receiver, \textbf{Prec.}: Precision, \textbf{Rec.}: Recall, \textbf{F-1}: F-1 Score \textbf{MRR:} Mean Reciprocal Rank, \textbf{MAE} Mean Absolute Error, \textbf{Syn.:} Synthetic}
      \\
      \multicolumn{8}{l}{*\textbf{NDCG}: Normalized discounted cumulative gain, \textbf{MAP :} Mean Average Precision}
    \end{tabular}
    }
  \end{adjustbox}
\end{table*}

\subsection{Probabilistic Method}
\label{subsec:probabilistic_approach}

A probabilistic approach for anomaly detection relies on the application of probabilistic models to model neighborhood relationships and patterns in dynamic graphs. Anomalies are determined based on a significant deviation from the model, considering a given threshold. This approach allows for the computation of p-values (or false positive rates) for their detection\cite{043_F-FADE_chang2021f}. However, this may require a complex optimization process to traverse a large graph dataset. Recent probabilistic methods include PENminer \cite{0103_belth2020_PENminer}, F-FADE \cite{043_F-FADE_chang2021f}, MIDAS \cite{040_bhatia2022_MIDAS_latest}, AnoEDGE \cite{056_bhatia2023_AnoEdge_sketch_based}, and several others.

Ranshous et al. \cite{111_ranshous2016scalable_CM-Sketch} introduced CM-Sketch, one of the earliest approaches for outlier detection in edge streams. CM-Sketch first considers both the global and local structural properties of the graph. It then utilizes the Count-Min sketch data structure to approximate these properties and provides probabilistic error bounds on their edge outlier scoring functions.

In 2020, Belth et al. \cite{0103_belth2020_PENminer} introduced PENminer, an anomaly detection approach for edge streams. PENminer focuses on exploring the persistence of activity snippets within evolving networks, which are essentially short sequences of recurring edge updates. Notably, PENminer is designed for both offline and streaming algorithms. 
The offline version leverages the measure to analyze time-stamped sequences of edges from historical data, while the online version, called sPENminer, calculates the measure incrementally for real-time analysis of edge streams. However, it is worth noting that PENminer is not equipped to detect subgraph and graph-level anomalies.

Chang et al. \cite{043_F-FADE_chang2021f} introduced F-FADE, a frequency factorization approach for AD in dynamic edge streams, which aims to detect anomalous edge streams by factorizing the frequency of the patterns. F-FADE \cite{043_F-FADE_chang2021f} discovers patterns by estimating the maximum likelihood rule of observed instances for each incoming interaction. It can effectively detect anomalies but requires a considerable amount of time and is computationally expensive. 

Bhatia et al. \cite{040_bhatia2022_MIDAS_latest} proposed a MIDAS probabilistic approach for detecting microcluster anomalies within edge streams. The algorithm employs count-min sketches (CMS) to compute the occurrence frequency of edges at each timestamp and subsequently utilizes the chi-squared test to assess the extent of deviation from typical edges, generating anomaly scores. Higher scores indicate the presence of anomalous patterns. Furthermore, the MIDAS algorithm maintains a stable level of memory utilization and a steady temporal complexity per edge. This method provides theoretical limitations on the chance of false positives, a characteristic that is not present in other probabilistic methods for anomaly detection in streaming. MIDAS \cite{040_bhatia2022_MIDAS_latest} also presents two distinct variants: Midas-R, which incorporates temporal and spatial relations, and Midas-F, which enhances precision by selectively filtering out anomalous edges. The MIDAS algorithm is one of the more recent dynamic edge stream anomaly detectors, and it requires constant memory and has a constant time complexity, which makes it scalable.

\textbf{MIDAS (2020)} vs. \textbf{F-FADE}: In a comparison between MIDAS and F-FADE, it is evident that MIDAS demonstrates greater scalability and computational efficiency when contrasted with F-FADE. However, MIDAS does have a notable limitation as it fails to track community structures, thus making it challenging to distinguish between various patterns. This particular limitation has been addressed in recent methods MSTREAM \cite{040b_bhatia2021_MSTREAM} and AnoEDGE \cite{056_bhatia2023_AnoEdge_sketch_based} by Bhatia et al., both of which are advancements on the MIDAS-R \cite{040_bhatia2022_MIDAS_latest} algorithm, by expanding the CMS to retain past dependencies and also implementing a higher-order sketch data structure to retain dense subgraph structures. 

\textbf{Experimental results} on three real-world dynamic graph datasets — DARPA \cite{d3_lippmann2000_DARPA} (network IP-IP traffics), CTU-13 (botnet traffic data), and UNSW-NB15 (a hybrid of real normal activities and synthetic attacks) — indicate that MIDAS-R \cite{040_bhatia2022_MIDAS_latest} provides an ROC-AUC scores of (\textbf{0.9514, 0.9703, 0.8517}) respectively, while F-FADE \cite{043_F-FADE_chang2021f} shows scores of (0.8451, 0.8028, 0.6858) on the respective datasets. Whereas, SEDANSPOT \cite{101_eswaran2018_SedanSpot} gives scores of (0.6442, 0.6397, 0.7575), and PENminer \cite{0103_belth2020_PENminer} provides scores of (0.8267, 0.6041, 0.7028). Compared to state-of-the-art methods, the MIDAS-R process evolves faster in constant time and memory, providing up to a $62\%$ higher ROC-AUC than state-of-the-art approaches.

MSTREAM \cite{040b_bhatia2021_MSTREAM} is a real-time streaming framework for detecting group anomalies in multi-aspect data. The goal is to detect anomalies, considering the similarity in categorical and real-valued attributes. MSTREAM utilizes the locality-sensitive hash functions \cite{040b_i_charikar2002_locality_sensitive_hash} to hash an incoming similar edge tuple into a fixed similar bucket, and then a temporal scoring function is applied to identify anomalous activity. The major difference between MSTREAM \cite{040b_bhatia2021_MSTREAM} and MIDAS \cite{040_bhatia2022_MIDAS_latest} is that MIDAS is designed to detect anomalous edges, which are two-dimensional records consisting of source and destination node indexes. Therefore, it cannot be applied in the high-dimensional context of multi-aspect data. On the other hand, MSTREAM extends MIDAS by assigning an anomalous score to each record and detecting anomalous records in a streaming manner.

Bhatia et al. \cite{056_bhatia2023_AnoEdge_sketch_based} introduced four sketch-based algorithms for detecting edge and graph anomalies in constant time and memory: AnoEdge-G and AnoEdge-L for edges, and AnoGraph and AnoGraph-K for graphs. These sketch-based algorithms build on MIDAS \cite{040_bhatia2022_MIDAS_latest} by expanding the count-min sketch (CMS) data structure to a higher-order sketch. The higher-order sketch data structure has the property of preserving dense subgraph structures in dense submatrix form, which simplifies the task of identifying a dense subgraph in a large graph to locating a dense submatrix in a fixed-size matrix. To the best of our knowledge, AnoEDGE and AnoGRAPH \cite{056_bhatia2023_AnoEdge_sketch_based} are the current state-of-the-art streaming edge and graph anomaly detection methods.

\subsection{Deep Learning Methods}
\label{subsec:deepLearning}

Deep learning, a subset of machine learning consisting of multiple interconnected neural networks, has been applied to address anomaly detection tasks. Notable techniques include autoencoders \cite{AD6_gong2019_Deep_AutoEncoder_unsuper,AD7_zong2018_deep_AutoEncoder_Gaussian_Mix, AD8_zhou2017_Robust_deep_autoEncoder}, generative adversarial networks (GANs), and RNNs \cite{AD12_su2019_robust_RNN_Anomaly_Detection}. However, conventional deep learning frameworks face limitations in handling streaming data characterized by intricate topological structures. Some of these challenges are discussed in Subsection \ref{1.1_deep_learning_challenges}.

Deep learning-based graph learning techniques leverage classical deep learning models for graph representation learning. These models fall into two broad categories: those directly adapted from other domains and those re-designed to suit the specific requirements of graph data embedding. We have grouped deep learning-based dynamic graph methods into four categories: auto-encoders, graph embedding, deep graph neural networks (GNNs), and graph transformer models. For more information, refer to Table \ref{tab2_AD_approach}.

\subsubsection{AutoEncoders} Autoencoders are a class of neural network architectures commonly used for anomaly detection in dynamic graphs. They have the ability to learn and reconstruct input data, and deviations from this reconstruction can indicate anomalies. In the context of dynamic graphs, autoencoders demonstrate adaptability to rapid changes in data distributions and are effective at capturing relevant information from nodes and edges. In many cases, variational autoencoders (VAEs) are utilized due to their probabilistic modeling approach for dynamic graphs. Additionally, autoencoders offer adaptive training and feature learning, making them well-suited for monitoring evolving graph structures. 

Notable techniques that employ autoencoders include DynGEM \cite{115_goyal2018_DynGem}, DynGraph2Vec \cite{116_goyal2020_Dyngraph2vec}, H-VGRAE \cite{057_yang2020_H-VGRAE}, and DGAAD \cite{117_gao2022anomaly_DGAAD}. H-VGRAE \cite{057_yang2020_H-VGRAE} constructs a hierarchical model by combining a variational graph autoencoder with a recurrent neural network. DGAAD \cite{117_gao2022anomaly_DGAAD} introduces a deep graph autoencoder model designed to acquire dynamic node embedding vectors for each node within the network. Initially, DGAAD employs a time-sensitive random walk algorithm to extract node sequences from the dynamic graph. Subsequently, it utilizes an auto-encoding approach to generate high-dimensional representation vectors for the nodes. Finally, the anomaly detection process is carried out by evaluating the network embeddings in terms of their proximity to the cluster center and their respective anomaly scores. 

\textbf{Experimental results:} experimenting with real-world dynamic graph data, autoencoders have yielded comparable results. For instance, DynGEM \cite{115_goyal2018_DynGem} achieved a noteworthy average MAP (mean average precision) evaluation score of \textbf{1.279} for link prediction on the ENRON dataset, outperforming all graph factorization baselines. Similarly, DGAAD \cite{117_gao2022anomaly_DGAAD} demonstrated an AUC of \textbf{0.7304} (for a $1\%$ anomaly) and 0.7197 AUC (for a $10\%$ anomaly injected) in the node prediction task on the ENRON dataset \cite{d4_shetty2004_Enron_Email}. When compared to the Node2Vec \cite{grover2016_Node2Vec} baseline methods, this represents a $10\%$ improvement on average and a remarkable $21\%$ improvement compared to Spectral Clustering with the Laplacian matrix for node embedding \cite{117_gao2022anomaly_DGAAD}. H-VGRAE \cite{057_yang2020_H-VGRAE} also gave a comparable result of an AUC score of $0.8366$ on the UCI message dataset, and a $0.7820$ AUC score on the Github dataset, which is slightly better than AddGraph \cite{026_AddGraph_zheng2019addgraph} with an AUC of $0.8083$ and $0.7257$ and NetWalk \cite{059_NetWalk_yu2018netwalk} with $0.7758$ and $0.6567$ for the respective datasets.

\textbf{Limitations}: Despite the competitive performance of Autoencoders and Deep Learning methods for anomaly detection in dynamic graphs, it is important to acknowledge the challenges they face. Autoencoders, often known as black-box models, suffer from a limitation in interpretability. Furthermore, they are computationally expensive to train on large dynamic graphs. These models also encounter difficulties in adapting to varying graph structures. Additionally, when applied to large-scale streaming graphs, both Autoencoders and Deep Graph Learning models may encounter scalability issues.

Therefore, it is recommended to explore alternative approaches, specifically probabilistic, density-based, and distance-based methods, in dynamic graph learning and representations. These techniques can offer valuable interpretability insights, explainability, speed, and scalability that may address the limitations associated with Autoencoders and Deep Learning models in detecting anomalous patterns in dynamic graphs.

\subsubsection{Graph Embedding} Graph embedding methods for anomaly detection involve applying graph embedding techniques to detect anomalies or outliers in graph-structured data. These methods aim to represent the graph's nodes and edges as vectors in a low-dimensional space, making it easier to identify nodes that deviate from the expected patterns or exhibit unusual behaviors within the graph. We cover the concept of graph embedding in depth in Section \ref{sec3:GNN_intro}.

Several graph representation techniques, such as DeepWalk \cite{perozzi2014_DeepWalk}, Node2Vec \cite{grover2016_Node2Vec}, LINE \cite{tang2015_LINE}, and NetWalk \cite{059_NetWalk_yu2018netwalk} have demonstrated their capability in generating node representations and have been used as a baseline model for recent graph learning methods. DeepWalk \cite{perozzi2014_DeepWalk} is a technique for graph embedding that relies on random walks. It creates random walks of a specified length originating from a target node and employs a skip-gram-like approach to acquire embeddings for unattributed graphs. LINE \cite{tang2015_LINE} aims to maintain the similarity between nodes in the first order and the proximity between nodes in the second order. Node2Vec \cite{grover2016_Node2Vec}, on the other hand, incorporates both breadth-first traversal (BFS) and depth-first traversal (DFS) in the generation of random walks. Similar to DeepWalk, it also utilizes the skip-gram algorithm to learn node embeddings. The major difference between DeepWalk \cite{perozzi2014_DeepWalk} and Node2Vec \cite{grover2016_Node2Vec} is that DeepWalk relies on random walks and is suitable for homogeneous graphs, whereas Node2Vec offers more flexibility by allowing both breadth-first and depth-first random walks, making it adaptable to heterogeneous graphs.

In contrast to DeepWalk and Node2Vec, the Netwalk \cite{059_NetWalk_yu2018netwalk} algorithm proposed by Yu et al. in 2018 focuses on capturing evolving network dynamics and scoring edge abnormalities in dynamic graphs, making it distinct from both DeepWalk and Node2Vec. In detail, the NetWalk \cite{059_NetWalk_yu2018netwalk} approach uses a random walk-based encoder for generating node embeddings, incorporating clique embeddings, and utilizing a graph autoencoder for the embedding learning process. It further captures the evolving nature of the network through dynamic reservoir updates. Finally, it utilizes a dynamic clustering-based anomaly detection method to assess the abnormality of individual edges.

The most recent graph embedding technique is AGR-AD \cite{036_fang2023_AER_anonymous_edge} by Fang et al., a method for detecting anomalies within dynamic bipartite graphs in an inductive setting. Their approach tends to capture the characteristics of an edge without using identity information. See Table \ref{tab2_AD_approach} for more details and comparison with other frameworks.

\subsubsection{Deep Graph Learning Based Techniques}

Zheng et al. \cite{026_AddGraph_zheng2019addgraph} proposed AddGraph, a framework that combines Gated Recurrent Unites (GRUs)  with attention mechanisms \cite{10_luong2015_attention_paper}, and temporal graph convolutional networks (GCNs) to detect anomalies in dynamic graph data. AddGraph considers both node-level and edge-level information in the graph to capture temporal dependencies. The attention mechanism is utilized to highlight important nodes and edges during anomaly detection.  AddGraph \cite{026_AddGraph_zheng2019addgraph} captures the structural information from the dynamic graph in each time stamp and the relationships between nodes. It has two layers, the GCN layer and the GRU layer.
Similar to Equation ~\ref{update}, at every time step, the  GCN utilizes the hidden state representation  $h_u^{(t-1)}$ at $t-1$ to generate the current node embeddings. Subsequently, the GRU-layer employs these node embeddings and hidden states attentions to learn the current hidden state $h^{(t)}$, as explained in Section \ref{subsec:3.2_message_passing}. Once the hidden state $h_u^{(t)}$ for all nodes is obtained, the AddGraph algorithm assigns an anomaly score to each edge in the dynamic graph, considering the associated nodes.

Park et al. proposed CGC  \cite{044_park2022_CGC}, a novel deep graph clustering approach that leverages a contrastive learning framework. CGC \cite{044_park2022_CGC} is designed to learn both node embeddings and cluster assignments in an end-to-end manner. It differs from other deep clustering methods, such as autoencoders, because it utilizes a multi-level scheme to carefully choose positive and negative samples. This ensures that the samples accurately reflect the hierarchical community structures and network homophily in the graph.

In 2020, Eswaran et al., the authors of SpotLight \cite{046_eswaran2018spotlight} and SedanSpot \cite{101_eswaran2018_SedanSpot} propose HOLS \cite{048_eswaran2020_HOLS_higher-order-label} (higher-order label spreading) a graph semi-supervised learning (SSL) approach that focuses on leveraging higher-order network structures. Traditional SSL methods rely on the homophily of vertices in graph networks, where nearby vertices are likely to share the same label. However, these methods often overlook the varying strengths of connections between vertices and the importance of higher-order structures in determining labels.

Cai et al. \cite{031_StrGNN_cai2021structural} introduced StrGNN, a GNN-based model for detecting anomalous edges. StrGNN leverages the graph convolution (GCN) operation and sorting layer to extract the h-hop enclosing subgraph of edges at each snapshot and proposes a node labeling function to identify the role of each node in the subgraph. Subsequently, stacked GCN and GRU layers are used to capture the graph's spatial and temporal dependencies. Finally, the model is trained in two stages: pre-training the SGNN using a graph reconstruction task and fine-tuning the entire STGNN for anomaly detection.

You et al. \cite{083_you2022_Roland} proposed ROLAND, an extension of the static GNN architecture to dynamic graphs. The primary focus is on snapshot-based representations for dynamic graphs, in which nodes and edges arrive in batches. Given the static node embedding state $H_t = \{ H_t^{(1)}, \dots, H_t^{(L)} \}$, ROLAND views $H_t$ as the hierarchical node state at time $t$, where each $H^{(l)}$ captures multi-hop node neighbor information. The ROLAND update module dynamically and hierarchically updates node embeddings over time.

Zhang et al.\cite{050_zhang2023_PageLink} proposed PaGELink, a path-based GNN explanation for heterogeneous link prediction tasks that generates explanations with connection interpretability. PaGE-Link works on heterogeneous graphs and leverages edge-type information to generate better explanations by reducing the search space by magnitude from subgraph-finding to path-finding and scales linearly in the number of edges.

Recently, Yuan et al. \cite{042_yuan2023_motif} introduced MADG, a motif-level AD method for detecting unique subgraph patterns in dynamic graphs without explicitly labeling anomaly data. MADG \cite{042_yuan2023_motif} first uses the motif-augmented stacked GCN to capture the topological relationships between nodes and motif instances and figure out how they are represented in each snapshot. Then, the generated representations of graph snapshots are input into a temporal self-attention layer to capture the temporal evolution patterns. Also, SAD \cite{053_tian2023_SAD}, a recent method based on the semi-supervised AD technique for dynamic graphs, utilizes the statistical distribution of unlabeled samples as the reference for loss calculation.

\begin{figure*}  

\includegraphics[width=0.9\textwidth]{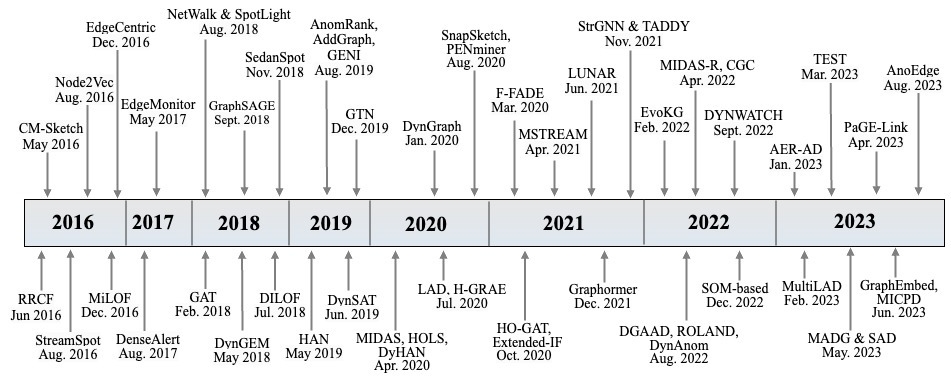}
  \caption{A timeline illustrating the chronological progression of Anomaly Detection (AD) methods in dynamic graphs from 2016 to 2023, as outlined in Table \ref{tab2_AD_approach}. The timeline reflects the publication years, including months, and denotes when each model was initially publicized. Note that the timeline may differ from the citation year if a paper was pre-published.}
  \label{fig4:AD_method_timeline}
\end{figure*}

\subsubsection{Graph Transformer}
 The Graph Transformer approach is a method for learning graph representations, which are commonly used in tasks like node classification, link prediction, and graph classification. The success of the transformer models \cite{10_luong2015_attention_paper, 11_yang2019cnn_selfattention,17_vaswani2017attention_is_all_you_need} in natural language processing (NLP) serve as inspiration for this adaptation to the graph domain. Graph transformer models are effective for capturing both local and global structural information within graphs, making them a valuable tool in graph-based machine learning tasks. Recent research continues to explore and develop new technique-based techniques. Recent transformer-based graph learning techniques  include GAT \cite{064_0velickovic2017_GAT}, GTN \cite{080_yun2019graph_Transformer_paper}, TADDY \cite{114_liu2021_TADDY}, and Graphomer \cite{063_ying2106_Graphomer_Do_transformers_Bad_Graph}.

GAT \cite{064_0velickovic2017_GAT} is a GNN-based model that uses the attention mechanism \cite{10_luong2015_attention_paper} on homogeneous graphs. HAN \cite{080b_wang2019_HAN} is a heterogeneous GNN model based on hierarchical attention, including node-level attention (to learn node importance) and semantic-level attention (to learn the importance of different meta-paths). HAN learns graph representation by transforming a heterogeneous graph into a homogeneous graph constructed by meta-paths. GTN \cite{080_yun2019graph_Transformer_paper} employs a transformer-based model to learn a new graph structure. This entails identifying valuable meta-paths and multi-hop connections between unconnected nodes within the original graphs.

Unlike GAT \cite{064_0velickovic2017_GAT} and HAN \cite{080b_wang2019_HAN}, where meta-paths are manually defined and graph neural networks are applied to meta-path graphs, GTN \cite{080_yun2019graph_Transformer_paper} learns meta-paths directly from the input graph data and performs graph convolutions on these learned meta-path graphs. This unique ability enables GTN to learn more useful meta-paths, leading to an effective node representation.


The most recent transformer-based methods for dynamic graph learning are Graphomer \cite{063_ying2106_Graphomer_Do_transformers_Bad_Graph} and TADDY \cite{114_liu2021_TADDY}. Ying et al. \cite{063_ying2106_Graphomer_Do_transformers_Bad_Graph} introduced Graphomer, a graph representation framework built upon the standard transformer model \cite{17_vaswani2017attention_is_all_you_need}. Graphomer incorporates various encoding techniques to learn graph information. First, it utilizes centrality encoding to capture node importance by leveraging degree centrality. Second, it applies spatial encoding to capture the structural relationships between nodes (i.e., edge encoding). The Graphomer centrality and spatial encoding are provided in Equations \ref{eqn4:centrality} and \ref{eqn5:spatial_enc}, respectively:

\begin{equation}
    \label{eqn4:centrality}
h_i^{(0)} = x_i + z_{\textit{deg}^{-}(v_i)}^{-} + z_{\textit{deg}^{+}(v_i)}^{+},
\end{equation}
where, $z^{-}, z^{+} \in \mathbb{R}^d$  represent the learnable embedding vectors associated with the in-degree $\textit{deg}^{-}(v_i)$ and the out-degree $\textit{deg}^{+}(v_i)$ of a directed graph, respectively.

For a spatial encoding technique, given any graph $G$, Graphomer \cite{063_ying2106_Graphomer_Do_transformers_Bad_Graph} proposes a function $\phi (v_i, v_j): V \times V \rightarrow \mathbb{R} $ to measure the spatial relation between $v_i$ and $v_j$ if two nodes are connected, else the output of $\phi = -1$:
\begin{equation}
\label{eqn5:spatial_enc}
    A_{ij} =  \frac{(h_i {W}_Q) (h_j W_K)}{\sqrt{d}} + b_{\phi (v_i, v_j)},
\end{equation}
where $A_{ij}$ is the $(i,j)$element of the Query-Key product matrix $A$ of the attention mechanism. This matrix, formed through the query-key product, is a foundational component of self-attention, enabling the model to selectively focus on various segments of the input sequence. Additionally, $b_{\phi (v_i, v_j)}$ is a learnable scalar indexed by $\phi (v_i, v_j)$ and shared across all layers.

TADDY \cite{114_liu2021_TADDY}, introduced by Liu et al., further expands the transformer-based model to a dynamic graph scenario. TADDY aims to detect anomalous edges within each timestamp while treating graph streams as a series of discrete snapshots. TADDY \cite{114_liu2021_TADDY} comprises four essential components: edge-based substructure sampling, spatial-temporal node encoding, a dynamic graph transformer, and the discriminative anomaly detector. This framework is trained end-to-end, enabling it to directly learn and output anomaly scores. The framework captures spatial-temporal contexts, integrates node information, and extracts knowledge from edges to calculate anomaly scores using a discriminative edge-scoring function. 

Other self-attention-based methods for dynamic graph learning include DySAT \cite{061_sankar2018_DySAT}, a dynamic self-attention network that computes node representations by simultaneously utilizing self-attention layers in two dimensions: structural neighborhood and temporal dynamics. DyHAN \cite{062_yang2020_DyHAN} is a dynamic heterogeneous graph embedding method that employs hierarchical attention to learn node embeddings. Additionally, HO-GAT \cite{060_huang2021_HO-GAT}, a hybrid-order graph attention method for detecting anomalous node and motif (or subgraph) instances within dynamically attributed graphs.

\subsection{Next-Generational Methods for Anomaly Detection in Dynamic Graphs}

Despite the recent success of statistical learning methods, probabilistic-based methods, matrix factorization, and deep learning-based methods, there are new emerging approaches for graph representation learning that have been explored in recent times. These emerging techniques include Quantum computing and Quantum Neural Networks (QNNs) \cite{20r_akbar2020towards_Quantum_Fruad_1, 21r_kukliansky2024network_AD_Quantum_2, 22r_rosenhahn2024_quantum_flowAD_3}, Federated learning for network traffic anomaly detection \cite{23r_pei2022personalized_federated_learning_1}, Reinforcement learning for anomaly detection in IoT \cite{24r_bikos2021_reinforcement_1}, and Graph Fourier Transforms (GFT) and spectral graph filtering for community-based anomaly detection. Specifically, we provide some quantum graph learning (QNNs) methods for AD in dynamic graphs.

\subsubsection{Quantum GNNs:} Quantum Graph Neural Networks (QGNNs) are types of neural networks that process graph-structured data and leverage the power of Quantum Computing (QC) to perform computations more efficiently than classical neural networks.

In 2023, Akbar et al. \cite{20r_akbar2020towards_Quantum_Fruad_1} proposed a Quantum Graph Neural Networks (QGNNs) model for financial fraud detection. The authors first constructed the graph representation for each transaction using PCA, and next, they encoded the representations into quantum states utilizing angle-encoding techniques. Furthermore, they utilized multi-layered Variational Quantum Circuits (VGC) to calibrate each quantum 6-qubits. The output of the VQC is subjected to average pooling, then fed through a linear layer, and finally to the output layer. QGNNs gave an AUC score of 0.85, which outperformed GraphSage with an AUC of 0.77 on a credit card fraud dataset with $284,807$ transactions. However, QGNNs challenges lie in the fact that it was not experimented with on real-world data, and model time complexity and scalability weren't recorded.

Most recently, in 2024, Kukliansky et al. \cite{21r_kukliansky2024network_AD_Quantum_2} proposed a Quantum Neural Networks (QNNs) approach for intrusion detection on noisy quantum machines. Experimenting on the real-world dynamic graph datasets KD-CUP99 and UNSW-NB15, their QNN approach gave an F1-score of $0.86$, outperforming classical neural network architectures like CNN (with a 0.636 F1-score) and MERA (with a 0.585).

In 2024, Rosenhahn et al. \cite{22r_rosenhahn2024_quantum_flowAD_3} proposed Quantum-based Normalizing Flows for anomaly detection. By comparing the distribution of quantum measurements, the authors computed a bijective mapping from the data samples to a normal distribution and then detected anomalies. The authors experimented with the Iris-Wine dataset and achieved an AUC score of 0.95 compared to classical Isolation Forest \cite{AD4_liu2008_isolation_Forest} with 0.92 and LOF \cite{AD1_2000_LOF} with 0.84. However, the authors did not experiment with real-world graph datasets, and the model running time is not recorded.

Despite the recent advancement in quantum computing and the comparative performance of QNN on graph anomaly detection, quantum-based algorithms are extremely complex due to quantum hardware challenges. Quantum data encoding on graph data requires advanced techniques; hence, this might not be a good technique for scalable streaming graphs.

\subsection{Timeline of Anomaly Detection Methods in Dynamic Graph}
The timeline presented in Figure \ref{fig4:AD_method_timeline} provides a chronological overview of anomaly detection (AD) methods in dynamic graphs spanning the years 2016 to 2023, as provided in Table \ref{tab2_AD_approach}. This timeline captures the advancement of AD models, showcasing their initial date of publication along with corresponding months. This will serve as a valuable visual representation, providing insights into the research progress made in dynamic graph anomaly detection techniques and highlighting the emergence of innovative methods over the specified timeframe.

\section{Dataset and Evaluation Metrics}
\label{sec6:dataset_metrics}

\subsection{Dataset}
Research studies on anomaly detection methods in dynamic graphs have mostly used real-world network data to quantify their performance level. However, a few others have also used synthetically generated data to simulate specific tasks.

In Table \ref{tab5:dataset_comparison}, we present an overview of dynamic graph datasets used in the current literature, along with links to the public repositories of those datasets. The most commonly used datasets include:
\subsubsection{UCI Message:} a directed and weighted network based on an online community of students at the University of California, Irvine. Each node represents a user, and each edge encodes a message interaction from one user to another, and the weight of each edge represents the number of characters sent in the message.
\subsubsection{Senate dataset:} is a social connection network between legislators during the
97rd-108th Congress \cite{d1_fowler2006legislative_Senate_dataset}. In this dataset, the 100th and 104th Congress networks
are recognized as the change points in many references. An edge is formed between two congresspersons if they cosponsored the same bill. Each bill corresponds to a snapshot and forms a clique of co-sponsors.
\subsubsection{Canadian bill voting network:} Extracted from the Canadian Parliament bill voting network. The Canadian Parliament consists of 338 Members of Parliament (MPs), each representing an electoral district, who are elected for four years and can be re-elected [11].
\subsubsection{Enron email data:}This data contains the email communication network between employees of the former US company Enron that has been made public by the US Department of Justice from January 2000 until April 2002 on a monthly level. 
\subsubsection{Ratings Data:} refers to 4-way tensors that include information about users, items, timestamps, and the associated ratings. These include the Yelp \cite{d2_Yelp_Data81:online}, Android, and YahooM datasets, as used in \cite{105_shin2017_DenseAlert}.

\subsubsection{Wikipedia Edit History:} \cite{d5_shin2016m_FastDense_Wiki_data} include the KoWiki and EnWiki datasets. This data consists of 3-way tensors that capture user interactions with Wikipedia articles. These tensors include information about users, pages (articles), and timestamps.

\subsubsection{Darpa:} \cite{d3_lippmann2000_DARPA} is a network traffic dataset simulating various intrusion behaviors. It contains 4.5M IP-IP communications (directed edges) taking place between 9,484 source IPs and 23,398 destination IPs (nodes) over 87.7K minutes.

\subsubsection{NycTaxi dataset:} \cite{d7_SNAP_dataset_Reddit} contains records of taxi ridership over a three-month duration, from November 2015 to January 2016, sourced from the New York City (NYC) Taxi Commission.
See Table \ref{tab5:dataset_comparison} for a comparison of commonly used dynamic datasets in the literature.


\begin{table*}[htbp]
    \caption{Summary of Commonly used Dynamic Graph Datasets in Literature}
    \label{tab5:dataset_comparison}
     \begin{adjustbox}{center}
  \resizebox{\textwidth}{!}{
    \begin{tabular}{l l l}
    \toprule
       \textbf{Dataset/Links } & \textbf{Application}& \textbf{ Description} \\
    \toprule
        \href{http://konect.cc/networks/opsahl-ucsocial/}{UCI Messages}  & Social Network  & Online  platform data of students at the University of California, Irvine. \\
        \href{https://snap.stanford.edu/data/soc-sign-bitcoin-alpha.html}{Bitcoin-Alpha/OTC}& Rating Networks & Rating networks collected from  Bitcoin platforms. \\

        \href{https://www.ourcommons.ca/en}{Canadian Bill-Voting} & Voting Network & Extracted form the Canadian Parliament bill voting network.\\

        \href{https://www.yelp.com/dataset}{Yelp} \cite{d2_Yelp_Data81:online} & Social Network & The dataset is a subset of Yelp's businesses, reviews, and user data. \\
        \href{https://ko-nlp.github.io/Korpora/en-docs/corpuslist/kowikitext.html}{koWiki, EnWiki} \cite{d5_shin2016m_FastDense_Wiki_data} & Wikipedia Edit History & Contain Wikipedia information such as articles, edits, and revisions \\
        \href{http://konect.cc/networks/youtube-u-growth/}{Youtube Favorite} \cite{d6_mislove2007_Youtube_social_network} & Social Network & Network data of YouTube users and their friendship connections. \\
        \href{https://www.ll.mit.edu/r-d/datasets/1998-darpa-intrusion-detection-evaluation-dataset}{DARPA} \cite{d3_lippmann2000_DARPA}
 &  Network Intrusion   & Contains network traffic logs simulating intrusion behaviors.\\
 \href{https://www.unb.ca/cic/datasets/nsl.html}{KDDCUP99} & Network Intrusion& {\parbox{10cm}{Based on the DARPA dataset, and it simulates network traffic including both normal and malicious activities.}} \\
 
CICIDS 2018 \cite{040b_bhatia2021_MSTREAM} &Network Intrusion& Network data generated at the Canadian Institute of Cybersecurity. \\
    UNSW-NB15 \cite{040b_bhatia2021_MSTREAM} & Network Intrusion  & {\parbox{10cm}{Hybrid of real normal activities and synthetic attack behaviors.  It contains nine types of attacks.}} \\
    CTU-13 \cite{d9_garcia2014_botnet_Data} & Network Intrusion &Botnet traffic dataset captured in the CTU University in 2011 \\
        \href{https://jmcauley.ucsd.edu/data/amazon/}{Android App rating} & Social Network & {\parbox{10cm}{A large crawl of product reviews from Amazon users.}} \\
        \href{https://networkrepository.com/dblp_coauthor.php}{DBLP Co-author} & Citation Network & {\parbox{10cm}{Graph dataset of authors from the DBLP computer science bibliography.}} \\
    \href{https://www.cs.cmu.edu/~enron/}{ENRON} \cite{d4_shetty2004_Enron_Email}& Communication Network & {\parbox{10cm}{Email communications between  Enron energy company  employees.}}  \\
    \href{https://networkrepository.com/email-dnc}{Email-DNC} & Communication Network& Network of emails in the 2016 USA, Democratic National Committee. \\

    Eu Email \cite{d7_SNAP_dataset_Reddit} & Communication Network & Timestamped edges
of emails sent within a European research institute \\
    \href{https://www.barracuda.com/}{BARRA} & Communication Network & {\parbox{10cm}{Collection of the email networks of the Barracuda Networks customers.}} \\
    \href{https://www.nyc.gov/site/tlc/about/tlc-trip-record-data.page}{NycTaxi} & Transportation Network &Contains records of taxi ridership over
a three-month  \\ 
    \href{https://www.kaggle.com/datasets/jackdaoud/bluebikes-in-boston}{Columbus Bike} & Transportation Network  & A bike trips dataset in the bike-share systems \\
    \href{https://www.kaggle.com/datasets/jackdaoud/bluebikes-in-boston}{Boston Bike} & Transportation Network & A bike trips dataset in the bike-share systems  \\
    \href{https://snap.stanford.edu/data/}{Reddit} \cite{d7_SNAP_dataset_Reddit} & Social Network & A collection of Reddit users post and  timestamped
references  \\
    Stackoverflow \cite{d7_SNAP_dataset_Reddit} & Social Network & Interactions among users on the
 Stackoverflow website \\
    TwitterWorldCup \cite{040_bhatia2022_MIDAS_latest} & Social Network & Contains 1.7M tweets for 2014 World Cup 2014 (June 12-July 13).  \\
    TwitterSecurity \cite{040_bhatia2022_MIDAS_latest} & Social Network & {\parbox{10cm}{Tweet with Department of Homeland Security keywords on terrorism.}}\\ 
    RTM \cite{d9_akoglu2008_RTM} & Social Network & Synthetic weighted time-evolving graph data on Kronecker products \\
    PolBlogs \cite{d10_adamic2005_PolBlogs} & Blog Network & Contains network of hyperlinks to blogs discussing the U.S. 2004  election. \\
    Pokec \cite{048_eswaran2020_HOLS_higher-order-label} & Friendship Network & Dataset of online friendship social network  in Slovakia \\
    AugCitation \cite{050_zhang2023_PageLink} &Citation Network & Constructed by augmenting the \href{https://paperswithcode.com/dataset/aminer}{AMiner citation network} \\
    \href{https://networkrepository.com/soc-digg.php}{Digg} & Social Network & {\parbox{10cm}{Collected from a news website digg.com where each node represents a user, and each edge represents a reply between two users.}}  \\
    
    \href{https://ogb.stanford.edu/}{OGB dataset} \cite{d11_hu2020open_OGB} & Graph Benchmark & {\parbox{10cm}{The Open Graph Benchmark (OGB) is a collection of realistic, large-scale, and diverse benchmark datasets for machine learning on graphs.}}  \\
    \href{https://networkrepository.com/tech-as-topology}{AS-Topology} & Network Data & Connection dataset collected from autonomous systems of the Internet. \\
    \href{https://www.kaggle.com/datasets/lakshmi25npathi/imdb-dataset-of-50k-movie-reviews}{IMDB} & Movie Review & {\parbox{10cm}{A movie dataset containing three types of nodes (movies (M), actors (A), and directors (D))}} \\
     \href{https://paperswithcode.com/dataset/acm}{ACM} & Citation Network & {\parbox{10cm}{ Contains dataset of papers published in KDD, SIGMOD, SIGCOMM, MobiCOMM, and VLDB.}} \\
     PPI \cite{d12_zitnik2017_PPI} &Protein Interaction & {\parbox{10cm}{ A protein-protein interaction (PPI) dataset that consists of graphs corresponding to different human tissues}} \\
      \href{https://www.alibaba.com/}{Alibaba} dataset & Social Network & Contains user behavior logs in the Alibaba.com e-commerce platform. \\
      \href{https://paperswithcode.com/dataset/webkb}{WebKB} &Social Network& Contains hyperlinked dataset of  877 web pages of four universities.  \\
      HEP-TH\cite{057_yang2020_H-VGRAE} &Citation Network& {\parbox{10cm}{citations of the papers in High Energy Physics Theory conference from 1993 to 2003}} \\
      AS-dataset \cite{116_goyal2020_Dyngraph2vec} & Network data &{\parbox{10cm}{AS (Autonomous Systems) a communication network of who-talks-to-whom from the BGP (Border Gateway Protocol) logs.}} \\
    
       \bottomrule
       
    \end{tabular}
    }
   \end{adjustbox}
\end{table*}
However, research on dynamic graphs is still relatively new and most cases of dynamic tasks tend to model real-world scenarios. Thus, it is a challenge to access real-world data, and this has hindered research and affects the reproducibility of experiments. One approach is to fall back to the generation of synthetic data. However, this may provide unrealistic scenarios with topological and attribute value limitations for graph-level tasks (node, edge, subgraph, graph).

\subsection{Evaluation Metrics}
\label{subsec:metric}
Commonly used metrics for evaluating the performance of anomaly detection techniques include accuracy, precision, recall, F1-score, AUC, Hit$@$n, MRR (mean reciprocal rank), MAE (mean absolute error), NDCG (normalized discounted cumulative gain), etc. In this section, we provide a detailed explanation of each of these metrics. Additionally, in Table \ref{tab6:metrics}, we present the mathematical definitions of these metrics.
\subsubsection{AUC-ROC} (Area Under the Receiver Operating Characteristic Curve) is a critical metric for binary classification model assessment. It quantifies the model's ability to distinguish between positive and negative classes at different classification thresholds. The \textbf{ROC} curve, which underlies AUC-ROC, displays the trade-off between true positive rate (TPR) and false positive rate (FPR) as the threshold varies. 
\subsubsection{Hit$@$n (Hits at n)} is a metric that reports the number of identified significant anomalies out of the top n most anomalous points.
\subsubsection{Hotelling T2 chart}, often referred to as the Hotelling's T-squared chart or $T^{\wedge} 2$ control chart, is a statistical quality control tool used in monitoring and detecting changes or shifts in multivariate data. It is an extension of the univariate control charts, such as the Shewhart chart, to handle multiple variables simultaneously.
\subsubsection{NMI (Normalized Mutual Information):} NMI is a metric used to measure the similarity between two clusterings of data. It quantifies the amount of information shared between two clusterings while accounting for the different numbers of clusters. A higher NMI value indicates a better similarity between the clusterings, with a maximum value of 1 indicating identical clusterings.

\subsubsection{ARI (Adjusted Rand Index):} ARI is another metric for assessing the agreement between two clusterings. It adjusts the Rand index to account for the expected value of random clustering. ARI yields a score between -1 and 1, where higher values indicate a better agreement between the clusterings, 0 represents a random agreement, and negative values indicate worse than random chance.
\subsubsection{NDCG (Normalized Discounted Cumulative Gain):} is a metric used to evaluate the quality of a ranked list of items, often in the context of information retrieval or recommendation systems. This metric considers both the predicted scores for the items and their graded relevance values, typically real-valued and non-negative ground truth scores. NDCG measures how well a ranking approach has performed in presenting the most relevant items at the top of the list. 
It is crucial to acknowledge that the metrics utilized in anomaly detection (AD) techniques for dynamic graphs extend beyond those listed in Table \ref{tab6:metrics}. A more specialized analysis is essential for comprehensive performance evaluation, as anomaly detection entails diverse requirements specific to applications and network topologies, as shown in Ma et al. \cite{024_sp_ma2021comprehensive_survey_AD_Deeplearning}.

\begin{table}[htbp]
  \caption{Commonly Used Evaluation Metrics in Literature}
  \label{tab6:metrics}
  \begin{tabular}{p{2cm} p{5cm}}
    \toprule
    \textbf{Metrics} & \textbf{Formula}\\
    \midrule
     Accuracy  & $\textit{Acc.} = \frac{(TN + TP)}{(TN + FN + FP + TP)}$ \\\\
     Precision & $\textit{Prec.} = \frac{TP}{(TP + FP)}$ \\ \\
     Recall/TPR &  $Rec. = \frac{TP}{(TP + FN)}$ \\\\ 
     F-1 score&  $F1 = 2 \times \frac{\textit{Recall } \times \textit{Precision}}{ (\textit{Recall } + \textit{Precision})}$  \\\\ 
     {\parbox{2cm}{Specificity\\(TNR)}}&  TNR = 1 - FP  \\\\ 
     AUC&  Area Under ROC curve  \\\\ 
     Hits$@ n$&  $\frac{\textit{ \# of detected anomalies at top n}}{n}$  \\\\ 
     MRR& {\parbox{4cm}{  $\textit{MRR} = \frac{1}{|U_{\textit{all}}|} \sum_{u=1}^{|U_{\textit{all}}|} RR (u)$ \\ \\ \\$RR(u) = \sum_{i\leq L} \frac{\textit{relevance}}{\textit{rank}_i}$ \\ \tiny \textbf{$|U_{all}|$:} total number of users  }} \\\\ 
    MAE &  $\textit{MAE} = \frac{1}{n} \sum_{i=1}^n \big\lvert y_i - y_i^{\textit{pred}}\big\rvert$ \\\\ 
    NMI & {\parbox{4cm}{  $\textit{NMI (Y, C)} = \frac{2 \times  I(Y; C)}{[ H(Y) + H(C)]}$\\ \\ \tiny \textbf{Y:} labels, \textbf{C:} clusters, \textbf{H(.):} Entropy, \textbf{I(Y;C):} mutual information b/w Y and C   }} \\\\
     NDCG&  $\textit{DCG} @ K = \sum_{i=1}^k \frac{r_i}{\log_2 (i+1)}$  \\
      \midrule
      \multicolumn{2}{l}{\footnotesize
    *\textbf{$r_i$} in $\textit{DCG} @ K$ is the graded relevance of the node at $i$} \\
    \multicolumn{2}{l}{\footnotesize
    *\textbf{RR (u)}: reciprocal rank of a user $u$ and the sum score for top $L$ }  \\
  \end{tabular}
\end{table}

\section{Challenges and Future Directions}
\label{sec7:future_works}
The field of graph representation learning and knowledge graphing (KG) has experienced rapid growth and attracted wide research attention over the past two decades. The recent increase in publications in prestigious Artificial Intelligence venues, as highlighted in Section \ref{sec5:Ad_approaches}, indicates that this trend is apparent not only in the static graph representation but also in dynamic graphs.

However, modeling graphs and detecting anomalous patterns in both discrete and continuously evolving graph structures remains a prominent concern, shaping potential future research directions. In this context, we examine common challenges and pinpoint potential directions for future research.

\subsection{Modeling Temporal Dynamics and Concept Drift}
Despite the increasing volume of research in graph representation and learning, anomaly detection in dynamic environments remains a challenging task, particularly in the context of modeling temporal-evolving networks and addressing concept drift.

\textbf{Temporal dynamics:} Researchers need to develop robust algorithms capable of adapting to evolving graph structures over time, such as \textit{inductive learning} algorithms. For instance, detecting self-propagating malware, such as \textit{polymorphic worms} that independently replicate and spread across computer networks, is a hot topic in cybersecurity and network analysis. Researchers could focus on creating more robust heuristic approaches with adaptive adjustment properties and incorporating behavioral analysis to detect and defend against such sophisticated attacks.

\textbf{Concept Drift:} Researchers need to investigate adaptive and hybrid algorithms for capturing dynamic graph structures by constantly adjusting to shifting data patterns and demonstrating real-time responsiveness. Additionally, there is a need to develop novel evaluation metrics that can measure the slight drift in the evolving nature of anomalies.

\subsection{Scalability}
Dynamic network modeling faces challenges when dealing with large-scale graph datasets that have a high volume of nodes and edges. To address these challenges, there is a need for robust and scalable algorithms for real-time anomaly detection.  The influx of data in streaming graphs makes modeling slow, less accurate, and computationally expensive. For example, approaches like \cite{042_yuan2023_motif,083_you2022_Roland,039_huang2020laplacian_LAD,039b_wang2017_Edge-Monitoring,115_goyal2018_DynGem,114_liu2021_TADDY} focus on learning graph patterns at each snapshot. However, this could lead to a high computational space as the network expands. Additionally, frequent snapshots could compromise the accurate modeling of temporal networks \cite{071_sp_skarding2021_foundations_dynamic_graph_survey}.

\textbf{Key questions on scalability}: How can dynamic graph models scale to adapt to continuous input stream length? What is the processing time per input node or edge compared to baseline approaches? 
Recent techniques like SedanSpot \cite{101_eswaran2018_SedanSpot}, which applied sub-processes like hashing, random walk, and sampling algorithms in sublinear time $ O(\log n)$, resulted in a large computation time. PENminer \cite{0103_belth2020_PENminer} and F-FADE \cite{043_F-FADE_chang2021f}, employing active pattern exploration and expensive frequency factorization operations, respectively, also resulted in large computation times. MIDAS-R \cite{040_bhatia2022_MIDAS_latest}, on the other hand, improved the complexity of streaming graph algorithms by applying the CMS (count-min sketch) \cite{111_ranshous2016scalable_CM-Sketch} hashing data structure in constant time and memory. However, it still struggles when the graph stream experiences exponential growth, exhibiting suboptimal performance in subgraph and graph-level anomaly detection tasks.

For future work, researchers could focus on improving continuous-time modeling by exploring distributed and parallel algorithms to address scalability issues associated with dynamic graphs and developing efficient data structures, as well as exploring algorithms like Count-Min Sketch with Conservative Update (CMSCU), FM Sketch (Flajolet-Martin Sketch), Lossy Counting, and other lightweight data structures that allow for more memory-efficient representations of streaming data.

\subsection{Multi-view Graph Anomaly Detection}
Multi-view anomaly detection refers to the task of identifying outliers in data represented as a graph with multiple views. Each view provides a distinct set of features associated with the nodes and edges of the graph. Researchers could focus on key concepts such as graph \textbf{heterogeneity} (i.e., the presence of diverse types of nodes, edges, or attributes within a graph), the integration of diverse views, ensuring proper alignment, and adapting multi-view anomaly detection algorithms into dynamic settings.

Exploring multi-view graph anomaly detection holds significant promise for future research directions in dynamic graph anomaly detection. Researchers could leverage multiple views of graphs to gain deeper insights into complex systems and uncover hidden anomalies that may not be apparent in single-view analyses \cite{039c2_huang2023laplacian_MultiLAD}. One key example is addressing graph heterogeneity, where the presence of different types of nodes, edges, or attributes within a graph leads to multi-layer anomalies \cite{039c1_xie2023_multi_view}. Furthermore, adapting these multi-view algorithms to dynamic settings could open up a new research direction for understanding temporal anomaly patterns and detecting evolving threats in real-time. Consequently, future research in this direction could lead to the development of more robust and versatile anomaly detection techniques capable of addressing the evolving challenges in dynamic graph data analysis.

Multi-view detection algorithms are applied in various domains, including social networks \cite{7_3a_MultiView_chen2023anomman}, time-series \cite{mf3_teng2017_multi_view_Time_series}, cybersecurity \cite{039c2_huang2023laplacian_MultiLAD}, and fraud detection \cite{7_3b_MultiView_Fraud_zhang2024mstan}, where a holistic understanding of complex relationships is crucial. Recent works, such as MultiLAD \cite{039c2_huang2023laplacian_MultiLAD, 039c1_xie2023_multi_view} and MSTREAM \cite{040b_bhatia2021_MSTREAM}, have targeted the detection of change point anomalies in multi-view graphs and group anomalies in multi-aspect data, respectively. However, this is an evolving research domain in dynamic graph learning.

\subsection{Multi-task Anomaly Detection}
Multi-task algorithms refer to models that simultaneously identify anomalous patterns across multiple related tasks or domains. Many approaches have focused on specific graph tasks, such as community detection \cite{042_yuan2023_motif, 025_alsentzer2020subgraph_SUBGNN}, node detection \cite{039_huang2020laplacian_LAD,053_tian2023_SAD, 120_goodge2022_LUNAR}, and edge- or link-level prediction \cite{040_bhatia2022_MIDAS_latest,043_F-FADE_chang2021f, 056_bhatia2023_AnoEdge_sketch_based}. However, in a complex and dynamic network, there may be two or more types of anomalies, posing a significant threat to critical infrastructure in cybersecurity. Attackers could potentially exploit the system with multiple kinds of attacks to bypass detection models.

To address this challenge, researchers could focus on developing dynamic fusion models that adaptively integrate information from multiple anomaly detection tasks. Additionally, future work could design algorithms that are task-aware, considering the unique properties within the network. Also, it's a good idea to encourage experts from different fields to work together on anomaly detection, dynamic graph theory, and improving multi-task anomaly detection systems.

\subsection{Graph Theoretical Foundation and Explainability}
Most existing anomaly detection algorithms for dynamic graphs are designed and evaluated through empirical experiments, lacking sufficient theoretical foundations to verify their reliability. Consequently, there is a tendency to overlook the explainability of the learned representations and detection results. For example, understanding which nodes, edge features, and adjacency matrices are most crucial in the graph or which edges or links predominantly influence the drift in network changes. Relying solely on anomaly scores may not be sufficient to conclusively determine if network traffic is anomalous or not.

Therefore, we believe that future work should focus on exploring the foundational knowledge of graph theory for modeling dynamic graph relationships and identifying patterns. This emphasis on the theoretical aspect of graphs will pave the way for new directions in model interpretability and advanced visual analytics. 

Additionally, future models could incorporate human-in-the-loop approaches to enhance the explainability and interpretability of dynamic graph algorithms. This holistic approach aims to bridge the gap between empirical evaluations and theoretical foundations, fostering a more comprehensive and reliable understanding of anomaly detection in dynamic graphs.

\subsection{Adversarial of Graph Models and Data Privacy}

The increase in graph-based research has also attracted a wide range of sophisticated attacks on graph-based models due to the availability and mining of big data from real-world networks \cite{032_sp_kyle2023survey_AD_IoT_Sensor_Networks, 2i_sp_ahmed2022_fake_news_survey}. Although several adversarial-resistant models have been developed recently, graph models are highly susceptible to structural adversarial attacks that can manipulate node or edge features to deceive the model into making incorrect detections \cite{024_sp_ma2021comprehensive_survey_AD_Deeplearning, 019_sp_kim2022_AD_with_GNN}. Graph-based models are particularly prone to evasion attacks involving subtle data modifications, data poisoning attacks, and data privacy concerns.

Future work could explore the development of techniques to mitigate such attacks and implement privacy-preserving techniques, such as differential privacy, to safeguard sensitive information in graph datasets.

\subsection{Fairness in Graph Anomaly Detection} 

Fairness refers to the equal and impartial treatment of individuals or societal groups by an AI system.
Addressing fairness in anomaly detection is crucial for ensuring equitable and effective models, especially in real-world applications. 
For instance, consider a fraud detection model that relies on historical data to predict anomalous transactions. If this data is biased, reflecting systemic discrimination against a specific demographic group, the model may flag potential transactions from that group. This could result in the unfair denial of legitimate transactions for individuals within that demographic or geographical segment.


Recent articles have explored these aspects in various contexts. Notable works include SRGNN by Zhang et al. \cite{14r_zhang2024_learning_Fairness_rebalancing} in 2024. This study addresses fairness issues related to sensitive node attributes by considering the impact of both low-degree and high-degree graph nodes in the GNN model for learning fair representations in decision-making. Furthermore, SRGNN employs adversarial learning to acquire fair representations through gradient normalization, ensuring the separation of each node’s representation from sensitive attribute information. Likewise, in 2024, Ling et al. \cite{15r_ling2024_fair_Feature} addressed the problem of fair feature selection for classification decision tasks in static graphs. The authors propose a fair causal feature selection algorithm called FairCFS. Specifically, FairCFS constructs a localized causal graph that identifies the Markov blankets of class and sensitive variables to block the transmission of sensitive information for selecting fair causal features.

In future research, it is important to integrate fairness into dynamic graph anomaly detection models. This involves not only considering the technical aspects of anomaly detection but also examining how these models may impact diverse communities and addressing potential biases. The development of algorithms that are fair, transparent, and unbiased will contribute to the responsible and ethical deployment of anomaly detection systems in various domains.

\subsection{Cost Sensitivity in Graph Anomaly Detection}

Cost sensitivity in anomaly detection involves accounting for varying costs associated with misclassifying anomalies or normal instances. This is particularly relevant in scenarios where the consequences of false positives or false negatives differ significantly. For instance, consider a credit card fraud detection system. The cost associated with a false negative (allowing fraud) is typically much higher than the cost of a false positive (blocking a legitimate transaction). Therefore, the anomaly detection models need to be cost-sensitive to prioritize minimizing false negatives, even if it means accepting a higher rate of false positives.

Several works have explored the concept of cost sensitivity, including the works of Zhang et al. \cite{16r_zhang2018_multiple_scale}. The authors propose a general target-resource framework involving multiple kinds of cost scales that minimize one kind of cost scale (called target cost scale) while controlling the others (called resource cost scales) in given resource budgets. Similarly, Huang et al. \cite{17r_huang2023regret_theory_multivariate} introduce a multivariate fusion prediction system that tackles the extraction of predictive information from multi-scale information systems. These approaches helped in assessing the data features more comprehensively and globally and highlighted the superiority degree between different samples.

In future research, it is important to integrate cost sensitivity into dynamic graph anomaly detection models. This involves designing algorithms that are not only accurate but also consider the economic implications of misclassifications. Developing models that are cost-sensitive will contribute to the practical and efficient deployment of anomaly detection systems in various domains.

\subsection{Further Research Directions}
Other open challenges and future work include, but are not limited to, the following:
\begin{itemize}

    \item Designing faster data streaming graph models, considering the tradeoff between speed and memory size.
    
    \item   Diversifying graph models for new applications, including environmental monitoring, medical data, and dynamic data streams.

    \item Exploring hybrid approaches (integration of deep learning models with streaming data structures)
    \item Addressing the imbalance problem in graph datasets
    \item Tackling other emerging research topics in graph representation learning
\end{itemize}

\section{Conclusion }

Due to the growing interest in research on graph representation learning and anomaly detection (AD) in dynamic graphs, we have conducted a comprehensive survey of existing AD methods in dynamic graphs. To the best of our knowledge, this is the most recent and holistic survey dedicated to anomaly detection in dynamic graphs, covering a wide range of modern techniques.

In Section \ref{sec2:backgroud}, we provided a concrete mathematical background and explained the different types of anomalies that can occur in both static and dynamic graphs. Section \ref{sec3:GNN_intro} discussed classical graph representation learning, specifically the GNN architecture, while Section \ref{sec4:dynamic_graph_rep} delved into dynamic graph representations. This set the stage for our survey on anomaly detection techniques in dynamic graphs. In Section \ref{sec5:Ad_approaches}, we reviewed and categorized current AD techniques, including (1) traditional machine learning (tree-based, density-based, and distance-based); (2) matrix factorization approaches; (3) probabilistic approaches; and (4) deep learning approaches. We presented a detailed summary and comparison of different anomaly detection techniques, current trends, and limitations. Furthermore, to aid future research advancement in this field, we presented a systematic timeline illustrating the chronological progression of all reviewed techniques. In Section \ref{sec6:dataset_metrics}, we also conducted a structured benchmarking of commonly used datasets (both real-world and synthetic data) and provided commonly used evaluation metrics for dynamic graph models. Finally, in Section \ref{sec7:future_works}, we highlighted potential research directions for future work based on the survey results.

We are optimistic that the rapid increase in research associated with dynamic graph learning will benefit numerous applications from diverse domains, and this survey provides a valuable contribution.

\begin{acks}
The authors wish to thank the College of Engineering, the Machine Intelligence and Data Science (MInDS) Center, and the Department of Computer Science at Tennessee Tech University for providing resources and funding to work on this project.
\end{acks}




\end{document}